\documentclass[10pt,twocolumn,letterpaper]{article}

\usepackage{cvpr}
\usepackage{times}
\usepackage{epsfig}
\usepackage{graphicx}
\usepackage{amsmath}
\usepackage{amssymb}

\usepackage{algorithm}
\usepackage{algorithmic}
\usepackage{caption}
\usepackage{subcaption}
\usepackage{multirow}
\usepackage{multicol}
\usepackage{pifont}

\usepackage[accsupp]{axessibility} 

\usepackage{hyperref}

\begin{document}

\title{ExFaceGAN: Exploring Identity Directions in GAN’s Learned Latent Space for Synthetic Identity Generation}


\author{Fadi Boutros$^{1}$, Marcel Klemt$^{1}$, Meiling Fang$^{1}$,  Arjan Kuijper$^{1,2}$, Naser Damer$^{1,2}$\\
$^{1}$Fraunhofer Institute for Computer Graphics Research IGD, Darmstadt, Germany\\
$^{2}$Department of Computer Science, TU Darmstadt,
Darmstadt, Germany\\
Email: fadi.boutros@igd.fraunhofer.de
}

\thispagestyle{empty}


\twocolumn[{
\maketitle
\begin{center}
\vspace{-6mm}
    \captionsetup{type=figure}
    \includegraphics[width=0.7\textwidth]{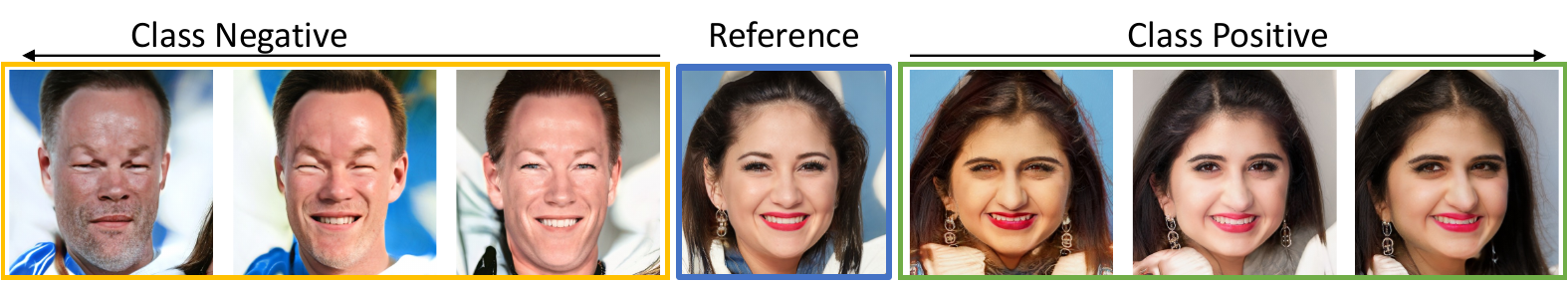}
    \captionof{figure}{Samples images generated by our ExFaceGAN applied to the learned latent space of unconditional StyleGAN-3}
\end{center}
}]



\begin{abstract}
   Deep generative models have recently presented impressive results in generating realistic face images of random synthetic identities.
   To generate multiple samples of a certain synthetic identity, previous works proposed to disentangle the latent space of GANs by incorporating additional supervision or regularization, enabling the manipulation of certain attributes.
   Others proposed to disentangle specific factors in unconditional pretrained GANs latent spaces to control their output, which also requires supervision by attribute classifiers.
   Moreover, these attributes are entangled in GAN's latent space, making it difficult to manipulate them without affecting the identity information.  
   We propose in this work a framework, ExFaceGAN, to disentangle identity information in
   pretrained GANs latent spaces, enabling the generation of multiple samples of any synthetic identity. 
   Given a reference latent code of any synthetic image and latent space of pretrained GAN, our ExFaceGAN learns an identity directional boundary that disentangles the latent space into two sub-spaces, with latent codes of samples that are either identity similar or dissimilar to a reference image. 
   By sampling from each side of the boundary, our ExFaceGAN can generate multiple samples of synthetic identity without the need for designing a dedicated architecture or supervision from attribute classifiers. 
   We demonstrate the generalizability and effectiveness of ExFaceGAN by integrating it into learned latent spaces of three SOTA GAN approaches.
   As an example of the practical benefit of our ExFaceGAN, we empirically prove that data generated by ExFaceGAN can be successfully used to train face recognition models (\url{https://github.com/fdbtrs/ExFaceGAN}). 
\end{abstract}
\vspace{-6mm}
\section{Introduction}
\label{sec:introduction}
\vspace{-1mm}
Recent advances in Deep Generative Models (DGM), especially Generative Adversarial Networks (GANs) \cite{GANs} and diffusion models \cite{DiffusionModel}, enabled the generation of photo-realistic face images.
The aim of DGMs is to learn the probability distribution of a certain training dataset, enabling the generation of completely new data points. Many DGMs also featured conditional image generation for structural and controllable outputs with a wide range of application scenarios, such as text-to-image generation \cite{DBLP:conf/cvpr/TewelSSW22}, photo-editing \cite{InterFaceGAN}, image-based virtual try-on \cite{DBLP:conf/wacv/FeleLPS22}, and synthetic-based face recognition \cite{USynthFace}.
While most of these application use cases require explicitly controllable image generation, others such as the development of FR using synthetic data require that the generated images are of discriminate synthetic identities and contain realistic variations that are not limited to small and predefined sets of attributes.

\begin{figure*}[ht!]
	\centering
    \includegraphics[width=0.90\textwidth]{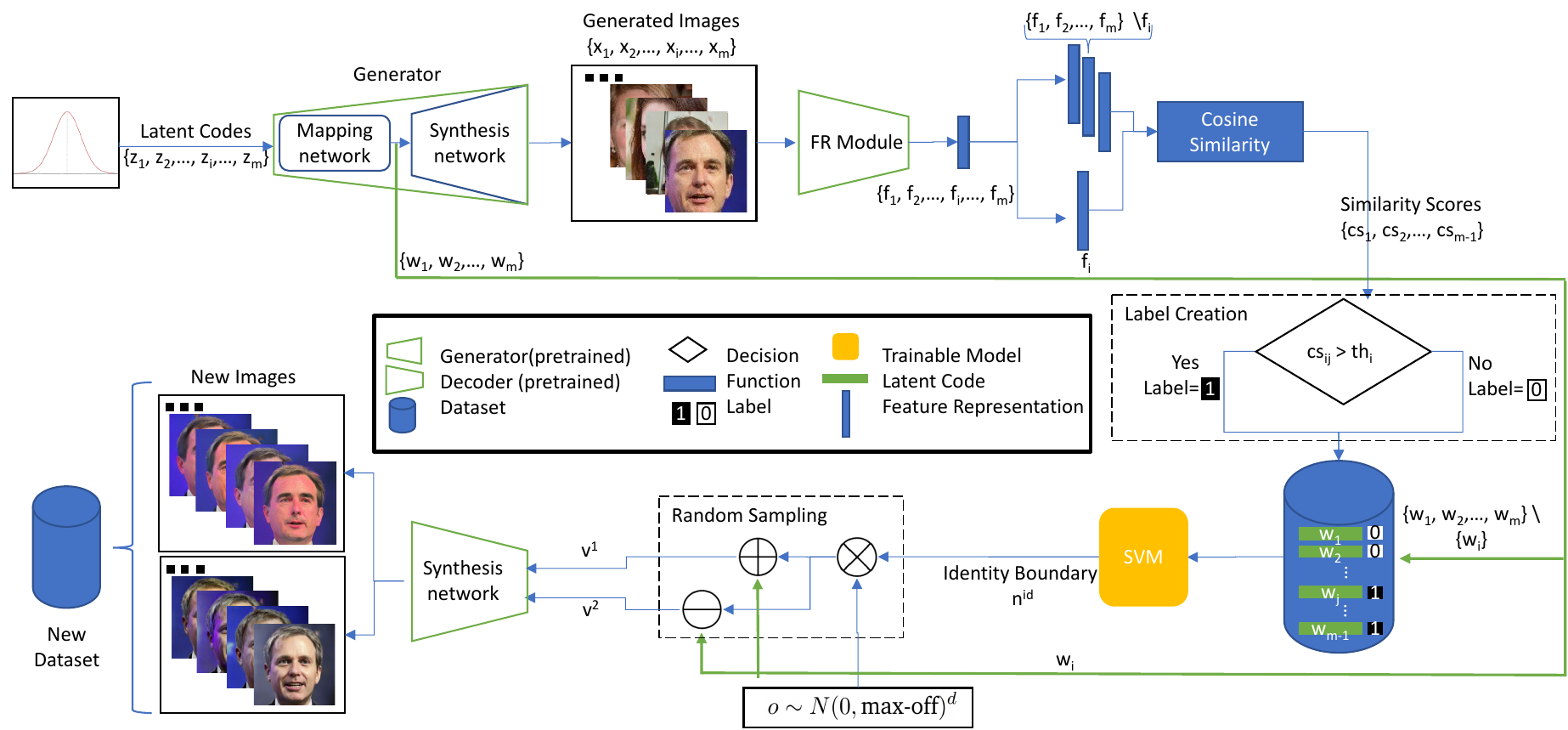}
	\caption{An overview of the proposed framework for discovering identity directions in learned latent spaces of GANS.
 }
 \vspace{-3mm}
\label{fig:id_discover_framework}
 \vspace{-2mm}
\end{figure*}

Recently, there was an increased interest in synthetic-based face recognition (FR) \cite{SFace,SynFace,USynthFace,digiface,IDiff-Face} driven by the increased legal and ethical concerns about the use, share, and management of real biometric data in FR development \cite{lirias3838501,DBLP:journals/ivc/BoutrosSFD23}. 
State-of-the-art (SOTA) synthetic-based FR \cite{SFace,SynFace,USynthFace,digiface} proposed either explicitly learning identity-discriminant feature representations \cite{USynthFace} or leaning multi-class classification problem \cite{SFace,SynFace,digiface}. In both learning strategies, these FR models relied on existing DGMs to generate multiple samples of synthetic identities.
 
Recent SOTA DGMs touching on the concept of generating multiple synthetic face images of synthetic identity with varying intra-class appearances can be grouped into two categories, controllable image generation by explicitly learning to generate face images with a predefined set of attributes \cite{DiscoFaceGAN, GAN_Control} and conditional image generation via manipulating learned latent space of unconditional DGMs \cite{SeFa, InterFaceGAN}. 
The approaches in the first category proposed to design and train conditional DGMs to explicitly generate synthetic images with a certain visual attribute, such as age, pose, illumination, expression, or combination of these attributes \cite{DiscoFaceGAN, GAN_Control}. 
SynFace \cite{SynFace} and UsynthFace \cite{USynthFace} utilized a controllable GAN, DiscoFaceGAN \cite{DiscoFaceGAN}, for their synthetic-based FR training. Each synthetic identity in their training datasets is formed by fixing the identity condition and randomizing the attribute conditions. 
However, the use of such controllable GAN for synthetic-based FR training suffers from two main drawbacks. First, the intra-class variations in the generated images are limited to a predefined set of attributes and do not necessarily reflect real-world variations. 
Second, extending these GAN models with additional attributes is extremely challenging as it requires designing and training a dedicated architecture for controlling additional attributes in the generated images.
SFace \cite{SFace} and IDNet \cite{Kolf_2023_CVPR} aimed at mitigating this challenge by training class-conditional GAN for class-labeled synthetic image generation. Images of each synthetic identity in SFace and IDNet are generated by fixing the class label and randomizing the generator's input latent code. However, the generated data suffer from low identity discrimination and the number of synthetic identities is limited to the number of classes in the training dataset. Unlike previous approaches, very recently IDiff-Face \cite{IDiff-Face} proposed  a latent diffusion model conditioned on identity contexts for synthetic identity generation, enabling the generation of multiple samples of synthetic identities with realistic variations.

The second category of image generation approaches proposed methods to manipulate the learned latent space of pretrained GANs, aiming at finding meaningful directions in latent space to produce a structural output generation \cite{SeFa, InterFaceGAN}. GANSpace \cite{GANspace} utilized Principal Component Analysis (PCA) applied on the feature space of pretrained GANs to create interpretable controls for image synthesis. Similar to GANSpace \cite{GANspace}, InterFaceGAN \cite{InterFaceGAN} proposed a framework for semantic face editing. InterFaceGAN trained a Support Vector Machine (SVM) on latent codes from a pretrained GAN latent space with labels from attribute classifiers to obtain a directional decision boundary for targeted attribute manipulations, enabling the generation of conditional images on visual attributes, e.g., adding eyeglasses or changing the pose. However, these approaches, GANSpace \cite{GANspace} and InterFaceGAN \cite{InterFaceGAN}, mainly relied on human labels or attribute classifiers for visual attribute manipulation without any restriction on identity information. 

In this paper, we propose a novel approach, ExFaceGAN, to discover identity directions in the learned latent space (or feature space) of a pretrained GAN generator, enabling the generation of multiple synthetic face images of a specific synthetic identity with realistic intra-class appearances.
Unlike previous works, the variations in the generated images by our approach are not limited to a predefined set of visual attributes and do not require human labeling or attribute classifiers to generate multiple samples of a synthetic identity.
In a nutshell, given a reference synthetic image of random identity with its latent code, our approach aims at disentangling the learned latent space with respect to the reference latent code into two sub-spaces, positive and negative subspace. The positive and negative sub-spaces contain latent codes of face images that are identity-similar and identity-dissimilar, respectively, to the reference images. These sub-spaces are corresponding to diverse face image transformations that maintain the identity information across the generated images in each discovered sub-space.
Thus, our approach can turn any pretrained unconditional GAN model into identity-conditional GAN without the need to change the model architectures or retrain the model. Also, we demonstrate that our ExFaceGAN can be integrated into attribute-conditional GAN models, enhancing the diversity in the generated images.
We additionally propose a sampling mechanism to control the inter-class variation and intra-class compactness of the generated data. 
An overview of our proposed ExFaceGAN is presented in Figure \ref{fig:id_discover_framework}.
Given a set of synthetic images, e.g. 5k images, and their corresponding latent codes, our ExFaceGAN can generate 10K discriminant synthetic identities with unlimited samples per identity. We empirically proved in this paper the identity discrimination in our ExFaceGAN-generated data. As an example of the practical benefit of ExFaceGAN, we demonstrated that the synthetically generated data by our ExFaceGAN can be successfully used to train FR models, outperforming all GAN-based FR models.

\vspace{-1mm}
\section{Methodology}
\label{sec:methodology}
\vspace{-1mm}
This section presents our framework for disentangling complex identity information in learned latent spaces of
pretrained identity-unconditional GAN.
The general idea of the proposed approach is to learn a directional boundary for each reference synthetic image that separates the latent space into two sub-spaces. The first latent subspace contains latent codes of synthetic images that share, to a certain degree, identity information with the reference image. The second latent subspace contains latent codes of images that are identity-dissimilar to the reference image. 
This disentangled latent space is then used to generate two new identity-specific sets of synthetic images.

Figure \ref{fig:id_discover_framework} illustrates the pipeline of our approach. We start by generating a set of images by sampling a set of latent codes from a Gaussian distribution and feeding it into a pretrained generator. One image, subsequently its latent code, is considered a reference for separating the latent spaces into the two sub-spaces.  
We label each synthetic image and its latent code with 1 (similar to the reference image) or 0 (dissimilar to the reference image) based on the similarity between the image and the reference one. The latent codes and their corresponding labels are then used to train an SVM to obtain an identity-separating boundary. By sampling latent codes from the two sides of the boundary, two identity-specific sets of images can be generated.
The pseudo-code of our algorithm is given in Algorithm \ref{algo:DIRGAN}.

\begin{algorithm}
\small
\caption{ExFaceGAN pipeline}
\begin{algorithmic} 
\STATE $M \leftarrow Mapping-network$
\STATE $S \leftarrow Synthesis-network$
\STATE $\phi \leftarrow FR-model$
\FOR{$i$ in $\{1,..., m\}$}
\STATE $z_i \leftarrow$ sample vector of size $(d)$ from $N(0,1)$
\STATE $w_i \leftarrow M(z_i)$
\STATE $x_i \leftarrow S(w_i)$
\STATE $f_i \leftarrow \phi(x_i)$
\ENDFOR
\FOR{$i$ in $\{1,..., m\}$}
\WHILE{$j \leq m$}
\IF{$j==i$}
\STATE continue
\ENDIF
\STATE $cs_{ij} \leftarrow \frac{f_{i} * f_{j}}{\Vert f_{i} \Vert \Vert f_{j} \Vert}$
\STATE $j \leftarrow j + 1$
\ENDWHILE
\STATE $th \leftarrow median(cs_i)$
\FOR{$j$ in $\{1,..., m-1\}$}
\STATE $labels_{j} \leftarrow 0$ \textbf{if} $cs_{ij} \leq th$ \textbf{else} $1$
\ENDFOR
\STATE $data \leftarrow W \setminus w_{i}$
\STATE $n_i^{id} \leftarrow SVM(data, labels)$
\ENDFOR
\STATE \textbf{Dataset creation}
\FOR{$i$ in $\{1,..., m\}$}
\WHILE{$a < num\_appearances$}
\STATE $o \leftarrow$ sample vector of size $(d)$ from $N(0,$max-off$)$
\STATE $v^1_a \leftarrow w_i \oplus o \otimes n_i^{id}$
\STATE $v^2_a \leftarrow w_i \ominus o \otimes n_i^{id}$
\STATE $q^1_a \leftarrow S(v^1_a)$
\STATE $q^2_a \leftarrow S(v^2_a)$
\STATE $a \leftarrow a + 1$
\ENDWHILE
\ENDFOR
\end{algorithmic}
\label{algo:DIRGAN}
\end{algorithm}
\vspace{-1mm}

\subsection{Disentangled Identity Representations in the learned Latent Space of GANs}
\label{sec:disentangle_id_representations}
Given an identity-unconditional pretrained GAN $G$ that is designed and trained to generate face images of random identities, we start by generating $m$ face images $X$. Specifically, $m$ latent codes $Z=\{z_1, z_2, ..., z_m\}$ are randomly sampled from a normal Gaussian distribution $ z_i \sim N(0,1)^d$. All latent codes are then processed by $G$ to output $m$ synthesized images $\{x_1, x_2, ..., x_m\}$, where $x_i$ is a synthetic image generated by $G$ from $z_i$. In StyleGAN-based architecture \cite{StyleGAN,StyleGAN2,StyleGAN3}, the $G$ consists of two networks, mapping and synthesis networks. The mapping network takes $z_i \in Z$ as an input and generates a style $w_i \in W$ (an intermediate latent code). $w_i$ is then passed to the synthesis network to generate images $x_i$. In StyleGAN-based approaches, the output of the mapping network ($W$) is commonly used as the latent space for GAN as it is learned to disentangle the inherent structure of the training data and thus, it contains more meaningful semantic information than $Z$ \cite{InterFaceGAN,StyleGAN}.
Each synthesis image with its latent code ($z_i$) and the intermediate latent code ($w_i$) forms a triplet $(z_i, w_i,x_i)$.
Formally, images are generated as follows:
\begin{equation}
    \left\{ x_i, w_i = G(z_i) \mid z_i \sim N(0,1)^d; i \in \{1, 2, ..., m\}\right\},
\end{equation}
where $x_i \in \mathbb{R}^{W \times H \times C}$, $z_i \in \mathbb{R}^{d}$ and $w_i \in \mathbb{R}^{d}$. 
Then, for each $w_i \in W$, we split $W \setminus w_i $ into two subsets, $W_{i1}$ and $W_{i2}$, where $W_{i1}$ contains latent codes of images that are identity similar to $x_i$  and ${W_{i2}}$ contains latent codes of images that are identity dissimilar to $x_i$.
The similarity between $x_i$ and each $x_j \mid j \in \{0,1...,m\} \setminus i$ is calculated in the embedding space between  $f_i \in \mathbb{R}^{l}$ and $f_j \in \mathbb{R}^{l}$ feature representations. Feature representations are extracted using a pretrained FR model $\phi$ as follows:
\begin{equation}
    \left\{ f_i = \phi (x_i) \mid i \in \{1, 2, ..., m\} \right\}.
    \label{eq:fr_inference}
\end{equation}

The similarity $cs_{ij}$ between $f_i$ and each $f_j$ is calculated using cosine similarity as defined in the following equation:
\begin{equation}
    cs_{ij} = \frac{f_{i} * f_{j}}{\Vert f_{i} \Vert \Vert f_{j} \Vert} \mid j \in \{1, 2, ..., m\} \setminus \{i\}.
    \label{eq:cos_sim_calculation}
\end{equation}

$f_j$ is considered to be similar to $f_i$ if $cs_{ij}$ is higher than a threshold $th_i$, and thus, the corresponding $w_j$ is in $W_{i1}$  (label of 1), otherwise $w_j$ is in $W_{i2}$  (label of 0).  
We consider the median cosine similarity of all $cs_{ij}$ as a threshold $th_i$. Formally $x_j$, its corresponding $f_j$, and latent codes $z_j$ and  $w_j$ is labeled as follows:
\begin{equation}
label_{j} = \begin{cases}
0 &\text{if $cs_{ij} \leq th_i$}\\
1 &\text{otherwise}
\end{cases}
\mid j \in \{1, 2, ..., m\} \setminus i.
\label{eq:threshold}
\end{equation}

$W \setminus w_i $ is split into two subspaces using a decision boundary obtained from an SVM. The SVM is trained on $W \setminus w_i $  and their corresponding binary labels. The normalized weights of the SVM define the direction  $n^{id} \in \mathbb{R}^{d}$ of the decision boundary, in which latent space $W \setminus w_i$ is split into two subspaces, i.e., latent codes located in the direction of $n^{id}$ and ones located in the opposite direction of $n^{id}$:
\vspace{-1mm}
\begin{equation}
\resizebox{0.9\linewidth}{!}{%
    $n^{id} = \text{SVM}(\{w_{j}\}, \{label_{j}\}) \mid j \in \{1, 2, ..., m\} \setminus i.$
    \label{eq:SVM_training}
}
\end{equation}

$n^{id}$ disentangles the latent space in relation to $w_{i}$. Images of the latent codes located in the direction of $n^{id}$ contain similar identity information to $x_i$, and thus, interpolating these codes with $w_i$ will lead to generating a set of images of that identity, i.e., class positive that are, to a large degree, similar to the reference image $x_i$.
Conversely, images of latent codes located in the opposite direction of $n^{id}$ are of dissimilar identities to the reference image $x_i$, and thus, interpolating these codes with $w_i$ will lead to generating another set of images of a new identity (dissimilar to $w_i$), i.e., class negative.
This procedure is repeated for each image $x_i \in X$, i.e., each $x_i$ is considered a reference once, resulting in $m$ decision boundaries. Each decision boundary can be used to sample two identity-specific sets of images.

\vspace{-1mm}
\subsection{Generation of Identity Specific Face Images}
\label{sec:generation_id_images}
\begin{figure}[h]
	\centering
    \includegraphics[width=0.80\linewidth]{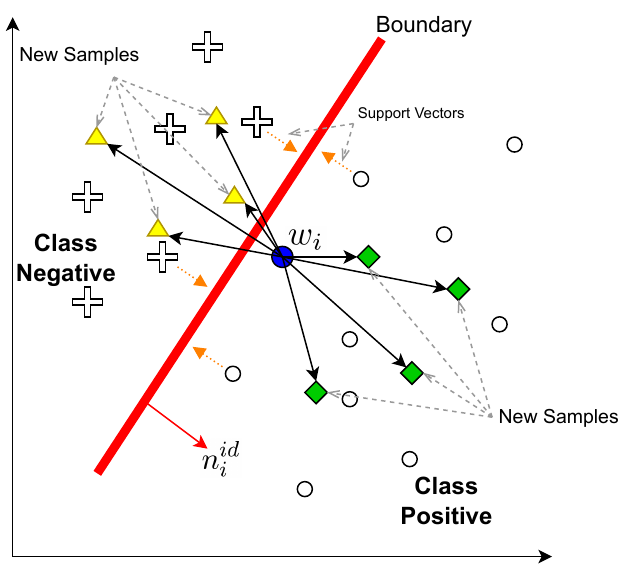}
    \vspace{-1mm}
	\caption{2D example of the proposed random sampling technique to obtain latent codes of positive and negative class. The boundary (red bar) belongs to $w_{i}$ and separates all data points into class positive and class negative. The red arrow $n^{id}$ indicates the normal directional vector of the boundary. The green rhombi are latent codes of class positive sampled from $w_{i}$ in the direction of $n^{id}$. The yellow triangles indicate latent codes of class negative sampled in the opposite direction of $n^{id}$.}
	\label{fig:random_sampling}
 \vspace{-4mm}
\end{figure}

\vspace{-1mm}
The decision boundary of SVM is used to obtain latent codes to synthesize new images of positive and negative identities.
To achieve this goal, we start by sampling an offset vector $o \in \mathbb{R}^{d}$ from $N(0,\text{max-off})^d$, where max-off is a hyperparameter that determines the maximal offset allowed in each dimension of $d$. An element-wise multiplication is performed between $o$ and $n^{id}$. Then, the resulting directional vector ($o \otimes n^{id}$) is added to $w_i$, resulting in a new latent code $v^1$.
A synthetic image generated from $v^1$ is of class positive, i.e., shared identity information with $x_i$.
Similarly, multiplication with negative $o$ will result in a latent code $v^2$ where the synthetic image using this latent code is of class negative, i.e., does not share identity information with $x_i$. By sampling different $o \otimes n^{id}$ and negative $o \otimes n^{id}$, we generate two sets of images of class positive and class negative as defined in the following:
\begin{align}
    v^1 = w_{i} \oplus o \otimes n^{id},  
    v^2 = w_{i} \ominus o \otimes n^{id},
\label{eq:random_sampling}
\end{align}
where $v^1, v^2 \in \mathbb{R}^{d}$ are the new latent codes sampled from class positive and negative sides, respectively. 
Images sampled from latent codes close to the boundary are assumed to be similar to image $x_{i}$. Increasing the max-off value of $v^1$ increases the appearance variations in the generated images while maintaining, to a very large degree, identity information of $x_{i}$. Conversely, increasing the max-off value of $v^2$ will result in synthetic images of a second identity (different than $x_i$). 
A simplified 2D example of our sampling technique is shown in Figure \ref{fig:random_sampling}.  $w_{i}$ refers to the latent code of the image $x_{i}$ for which the boundary $n^{id}$ was trained. The red arrow $n^{id}$ indicates the normal vector of the identity-separating boundary. The boundary separates all data points into class positive and class negative, i.e., points located in the positive and negative (opposite) direction of the decision boundary, respectively. 

\vspace{-2mm}
\section{Experimental Setup}
\label{sec:exp_setup}
\textbf{Identity discrimination evaluation metrics:} We evaluate the identity discrimination in our ExFaceGAN datasets using the following metrics: Equal Error Rate (EER),  FMR100  which is the lowest False None-match Rate for False Match Rate (FMR)  $\leq$ 1.0\% \cite{iso_metric}. Also, we report the Fisher Discriminant Ratio (FDR) \cite{poh2004study}.
FDR metric indicates the separability of the genuine and impostor distribution. To calculate the genuine and imposter comparison scores, we utilize the ArcFace model \cite{ArcFace} \footnote{ArcFace model architecture is ResNet50 \cite{ResNet} trained on CASIA-WebFace dataset \cite{CASIA}.} for feature extraction.

\textbf{Dataset generation:}
We applied our ExFaceGAN on three  generative models (official pretrained releases), StyleGAN2-ADA \cite{StyleGAN2Ada}, StyleGAN3 \cite{StyleGAN3}, and GAN-Control \cite{GAN_Control}, noted as ExFaceGAN(SG2), ExFaceGAN(SG3), and ExFaceGAN(Con), respectively. StyleGAN models are unconditional generative models, while GAN-control is an attribute conditional (age, expression, illumination, and pose ) model. We opt to further apply our approach to GAN-Control \cite{GAN_Control} to demonstrate the applicability of our ExFaceGAN to both, unconditional and attribute-conditional generative models.
For each of the considered models, ExFaceGAN datasets are created by randomly sampling 5k different latent codes from a normal Gaussian distribution. Then, 5k $w$ representations are obtained from each model mapping network. The $w$ is then processed by each of the  synthesis networks to produce 5k synthetic images. It should be noted that all experiments are conducted in the w-space and not in the z-space of GANs as w-space provides a more disentangled representation \cite{StyleGAN, InterFaceGAN} than z-space. 
All synthetic images are aligned and cropped using five landmarks similarity transformation as defined in \cite{ArcFace}. The landmarks are extracted using MTCNN \cite{MTCNN}. 
For each model, we train 5k SVMs to obtain identity-separation boundaries, following our algorithm described in \ref{sec:disentangle_id_representations}. Finally, ExFaceGAN datasets are generated as defined in Section \ref{sec:generation_id_images}.

\begin{figure}[ht!]
	\centering
    \begin{subfigure}[t]{0.49\textwidth}
        \centering
        \includegraphics[width=0.99\linewidth]{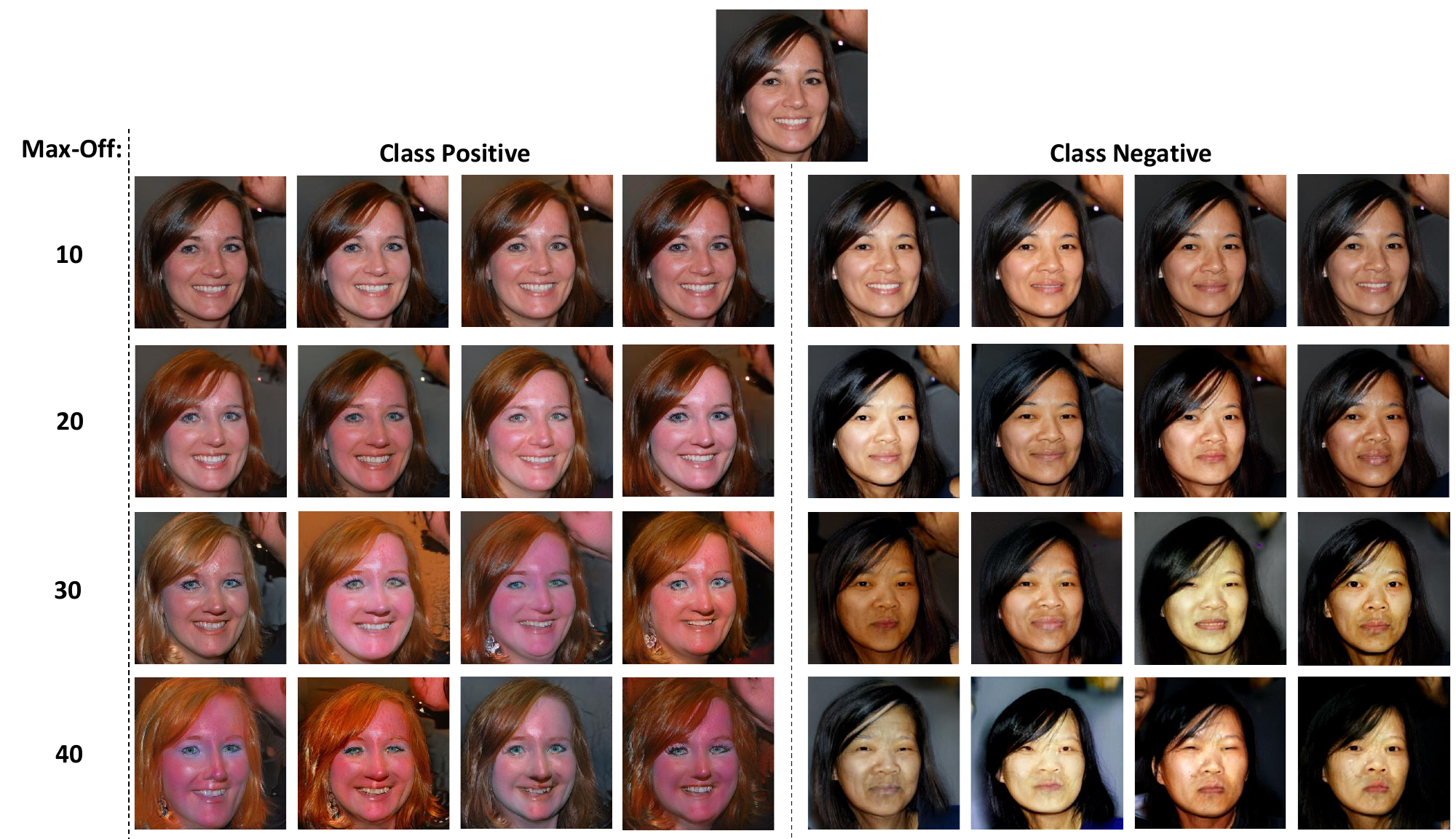}
        \caption{ExFaceGAN(SG2) }
	    \label{fig:SG2_intra_class_variation}
    \end{subfigure}
    \begin{subfigure}[t]{0.49\textwidth}
        \centering
        \includegraphics[width=0.99\linewidth]{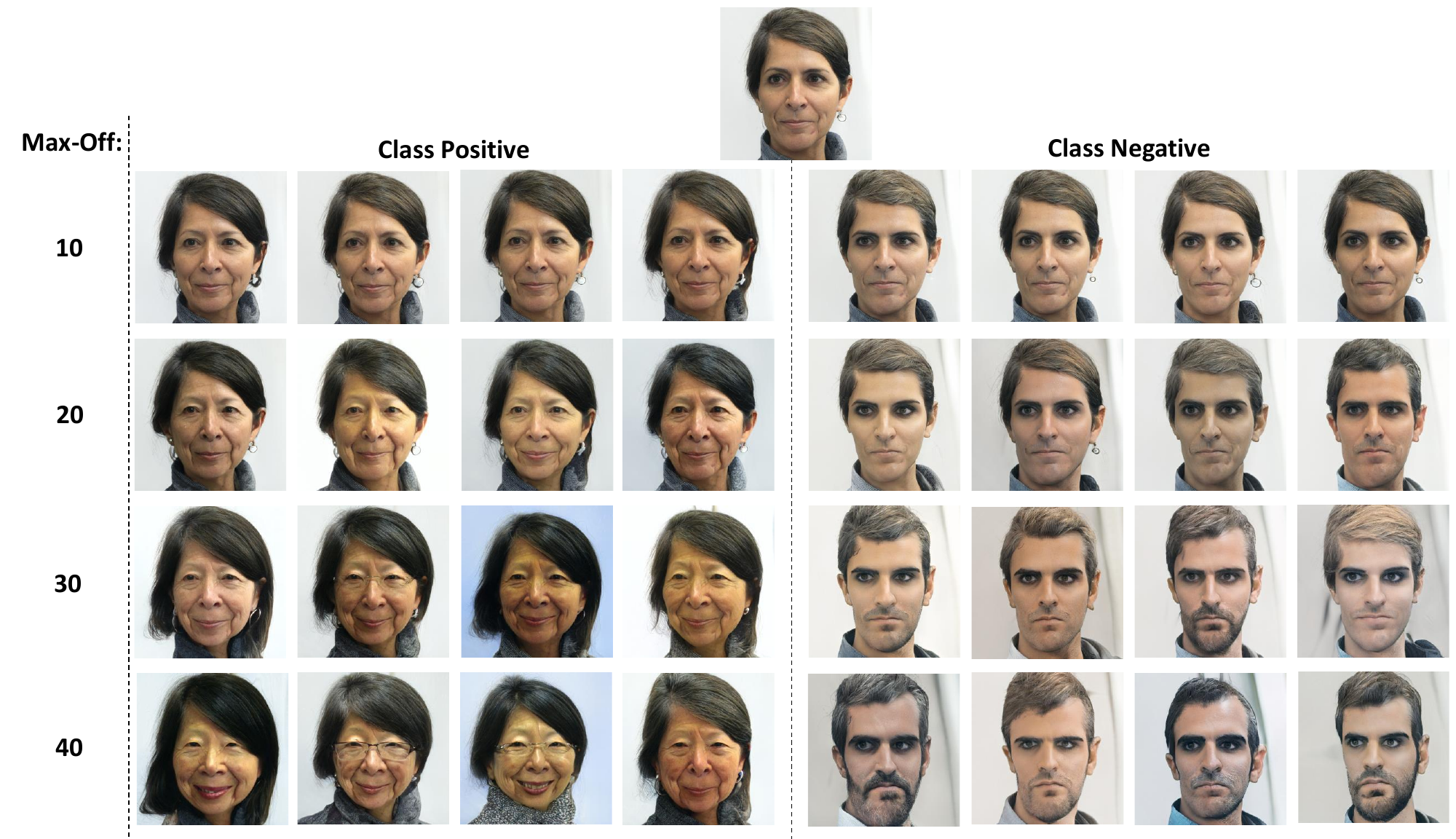}
        \caption{ExFaceGAN(SG3)}
	    \label{fig:SG3_intra_class_variation}
    \end{subfigure}
    \begin{subfigure}[t]{0.49\textwidth}
        \centering
        \includegraphics[width=0.99\linewidth]{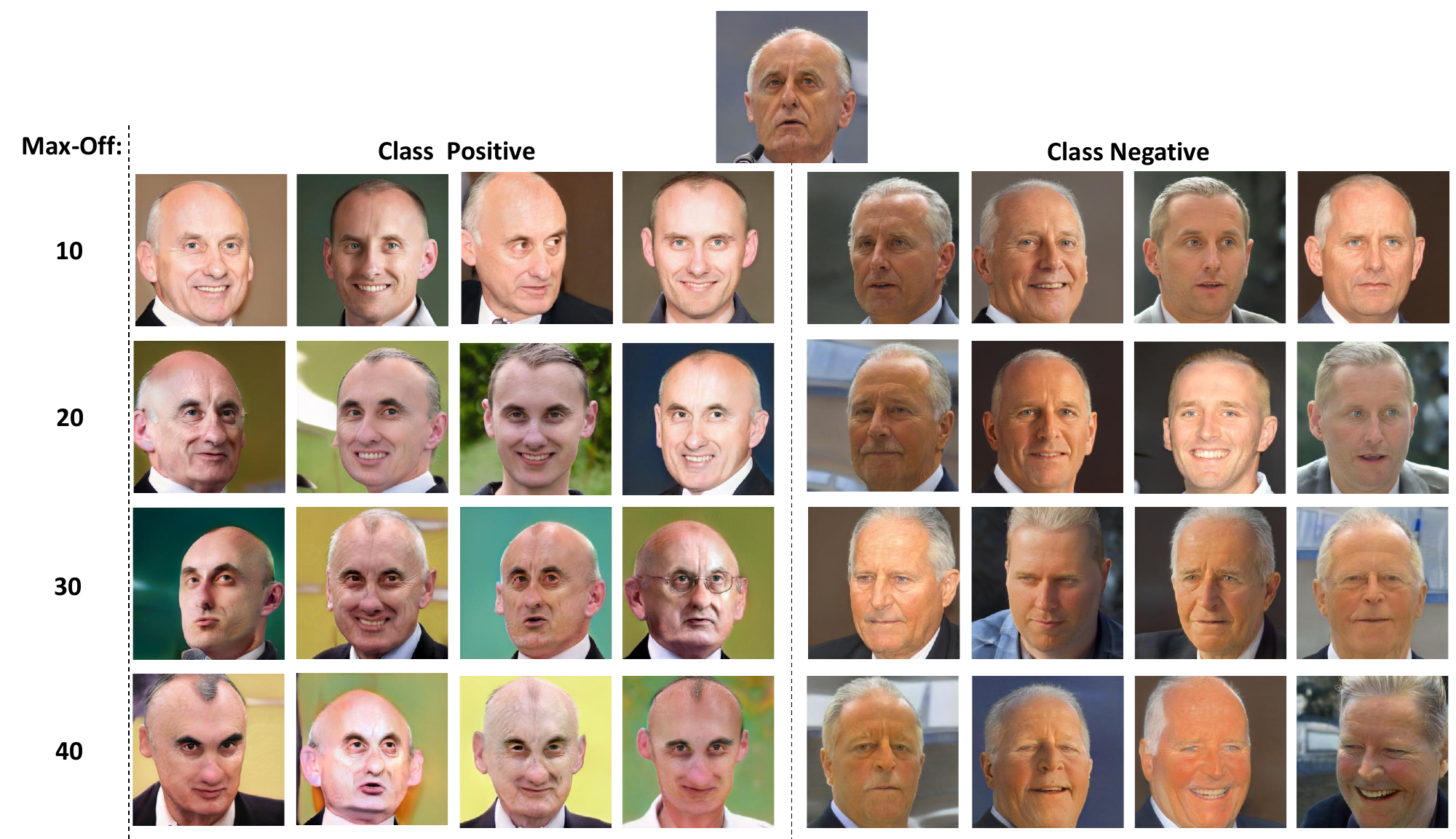}
        \caption{ExFaceGAN(Con) }
	    \label{fig:GAN_C_intra_class_variation}
    \end{subfigure}
	\caption{Samples from class positive and negative with different max-off values generated by our ExFaceGAN(SG2), ExFaceGAN(SG3), and ExFaceGAN(Con). The images on the top are the reference images.
 }
\label{fig:samples_dirgan}
\vspace{-4mm}
\end{figure}

\textbf{FR model training setup:}
We evaluate the verification accuracies of FR models trained on our ExFaceGAN datasets. Following the previous synthetic-based FR training approaches \cite{SynFace,SFace,USynthFace}, we use ResNet-50 network architecture \cite{ResNet} with ArcFace loss (margin penalty of 0.5 and scale term of 64 \cite{ArcFace}). During the training, a dropout of 0.4 is applied. 
Stochastic Gradient Descent (SGD) is employed as an optimizer with a learning rate of 0.1. The learning rate is divided by 10 at 22, 30, and 35 epochs \cite{ArcFace,ElasticFace}. The model is trained for 40 epochs with a batch size of 512.

\begin{figure}[ht!]
	\centering
    \begin{subfigure}[t]{0.23\textwidth}
        \centering
        \includegraphics[width=\linewidth]{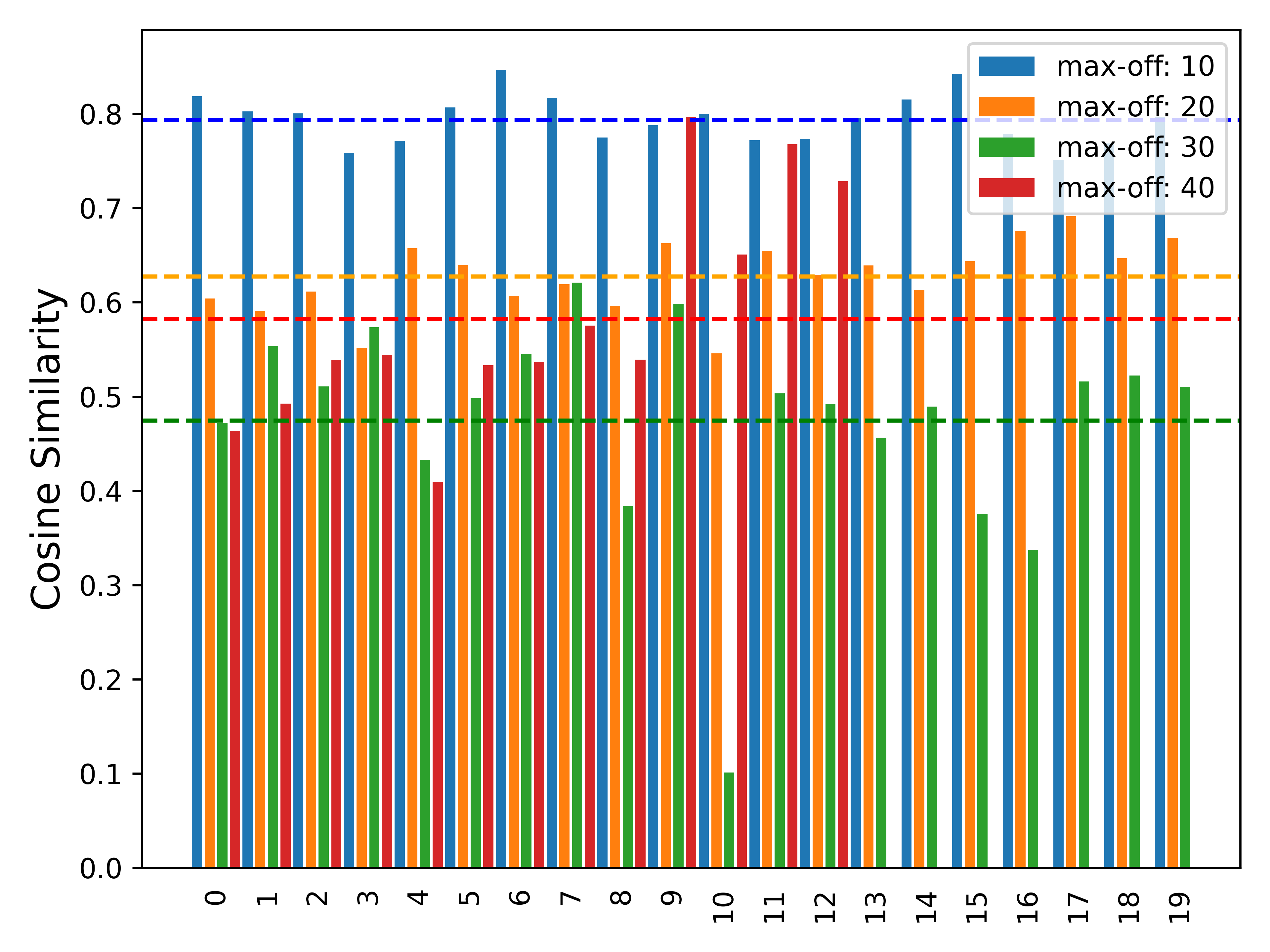}
        \caption{ExFaceGAN(SG2)}
        \label{fig:cos_sim_SG2_class1}
    \end{subfigure}
    \begin{subfigure}[t]{0.23\textwidth}
        \centering
        \includegraphics[width=\linewidth]{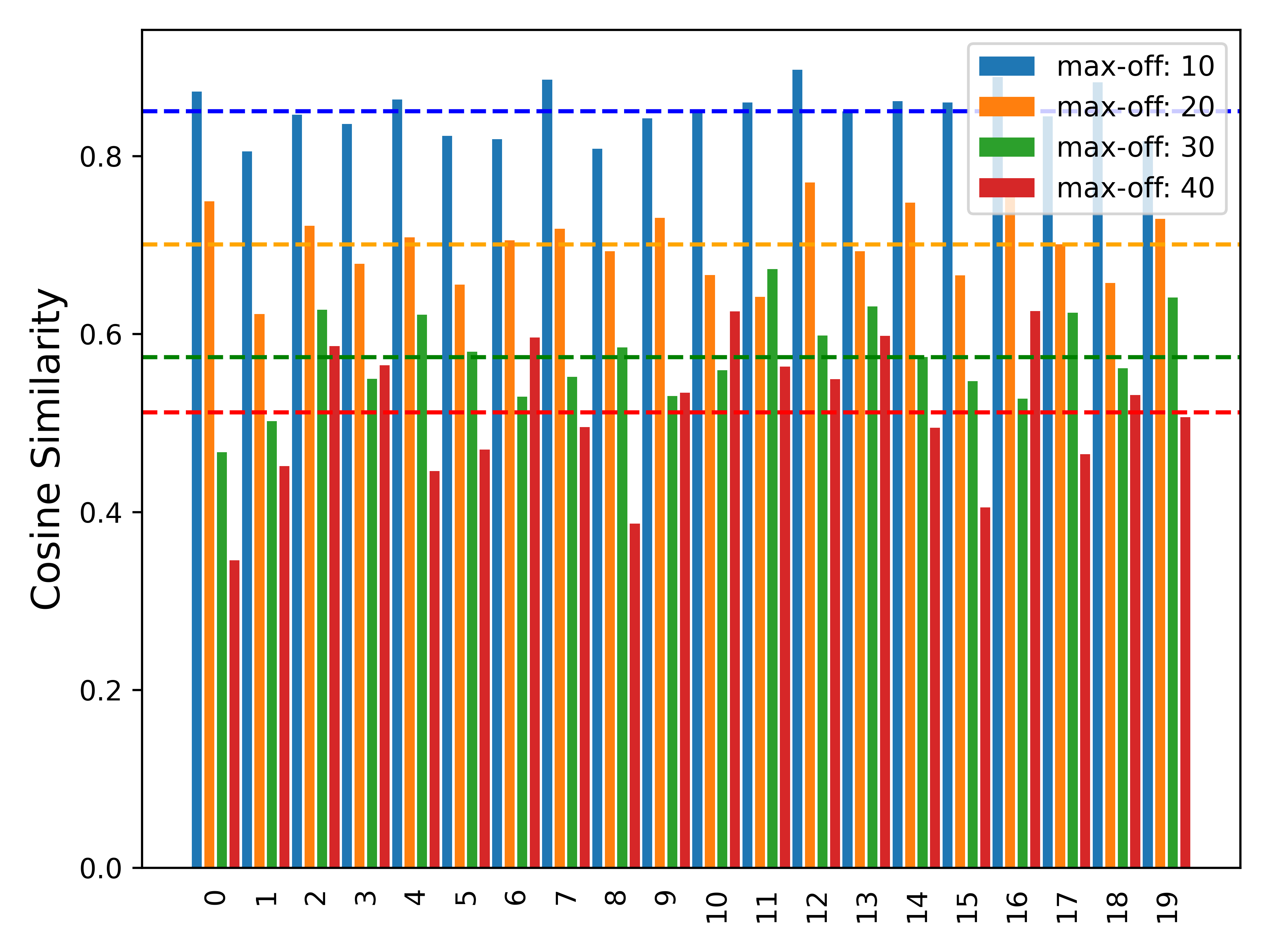}
        \caption{ExFaceGAN(SG3)}
        \label{fig:cos_sim_SG3_class1}
    \end{subfigure}
    \begin{subfigure}[t]{0.23\textwidth}
        \centering
        \includegraphics[width=\linewidth]{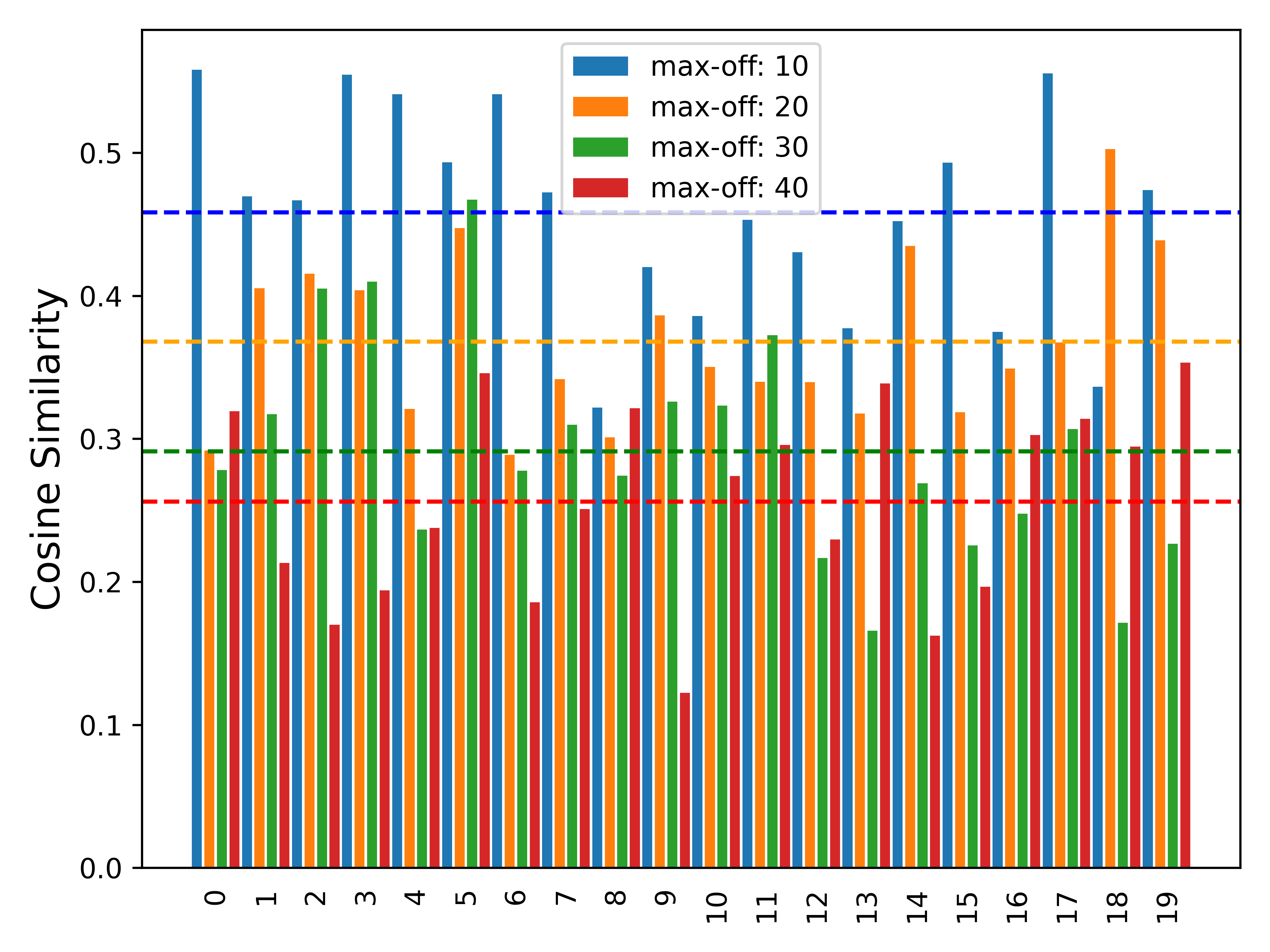}
        \caption{ExFaceGAN(Con)}
        \label{fig:cos_sim_GAN_C_class1}
    \end{subfigure}
    \begin{subfigure}[t]{0.23\textwidth}
        \centering
        \includegraphics[width=\linewidth]{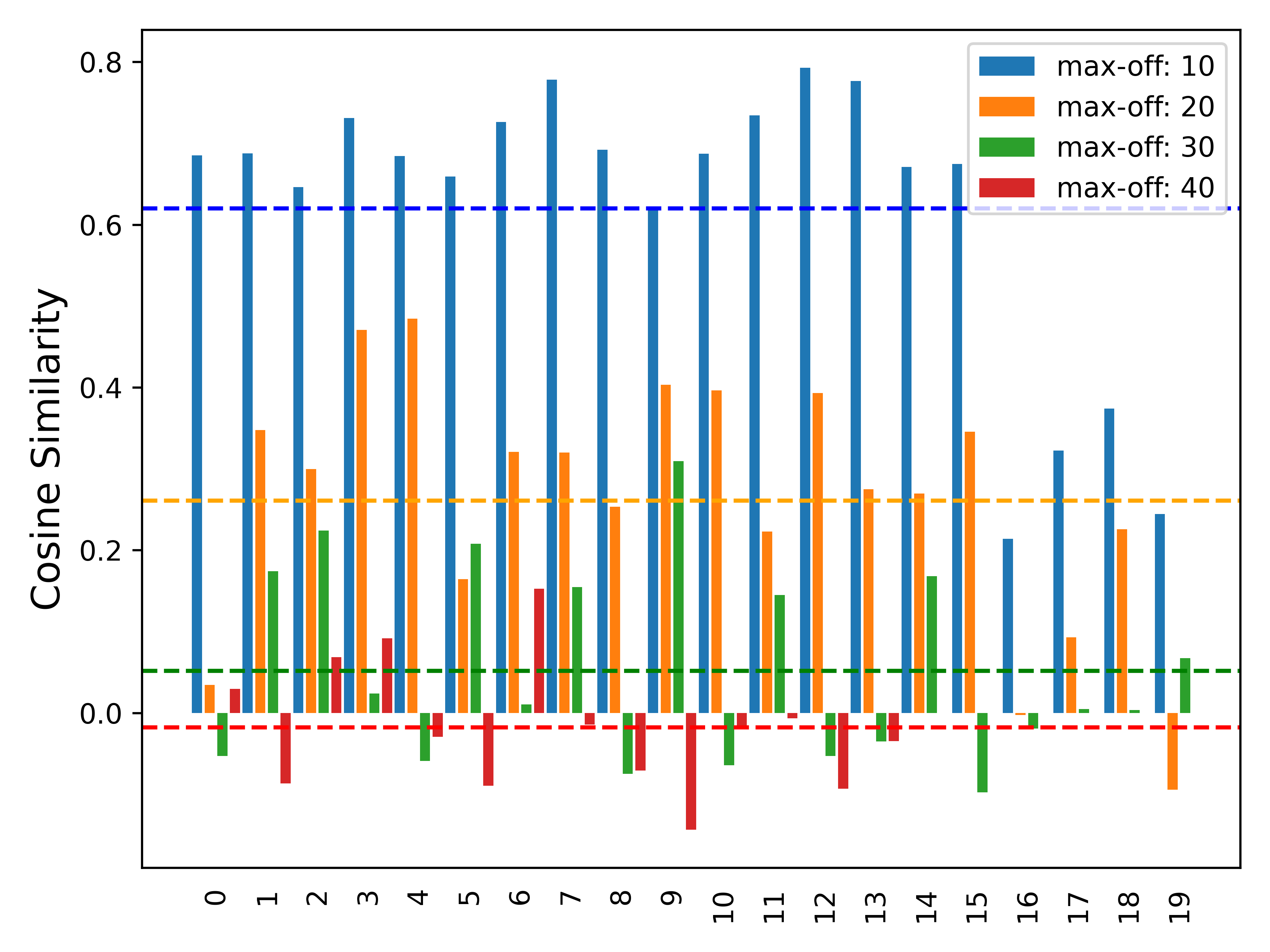}
        \caption{ExFaceGAN(SG2)}
        \label{fig:cos_sim_SG2_class2}
    \end{subfigure}
    \begin{subfigure}[t]{0.23\textwidth}
        \centering
        \includegraphics[width=\linewidth]{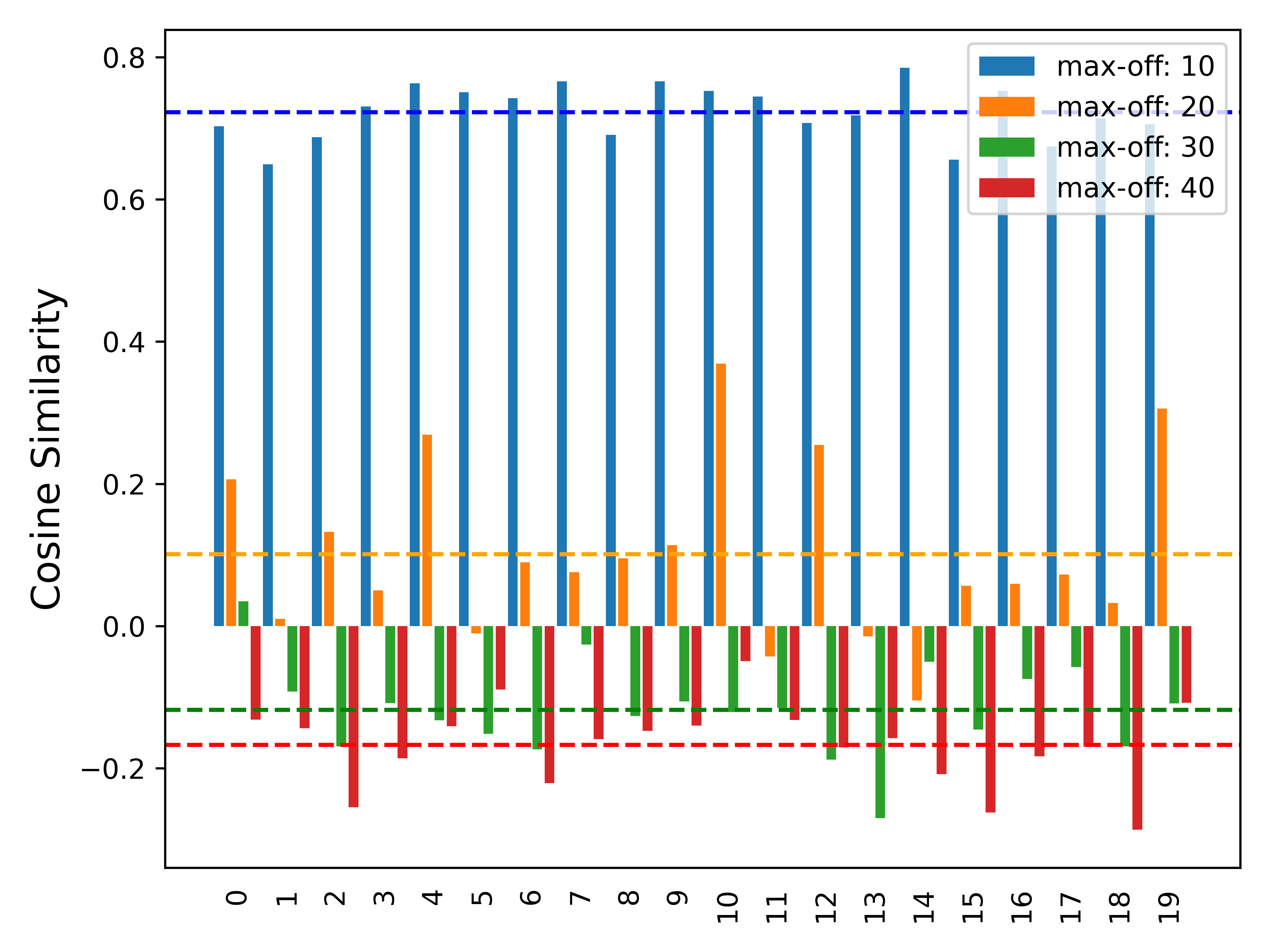}
        \caption{ExFaceGAN(SG3)}
        \label{fig:cos_sim_SG3_class2}
    \end{subfigure}
    \begin{subfigure}[t]{0.23\textwidth}
        \centering
        \includegraphics[width=\linewidth]{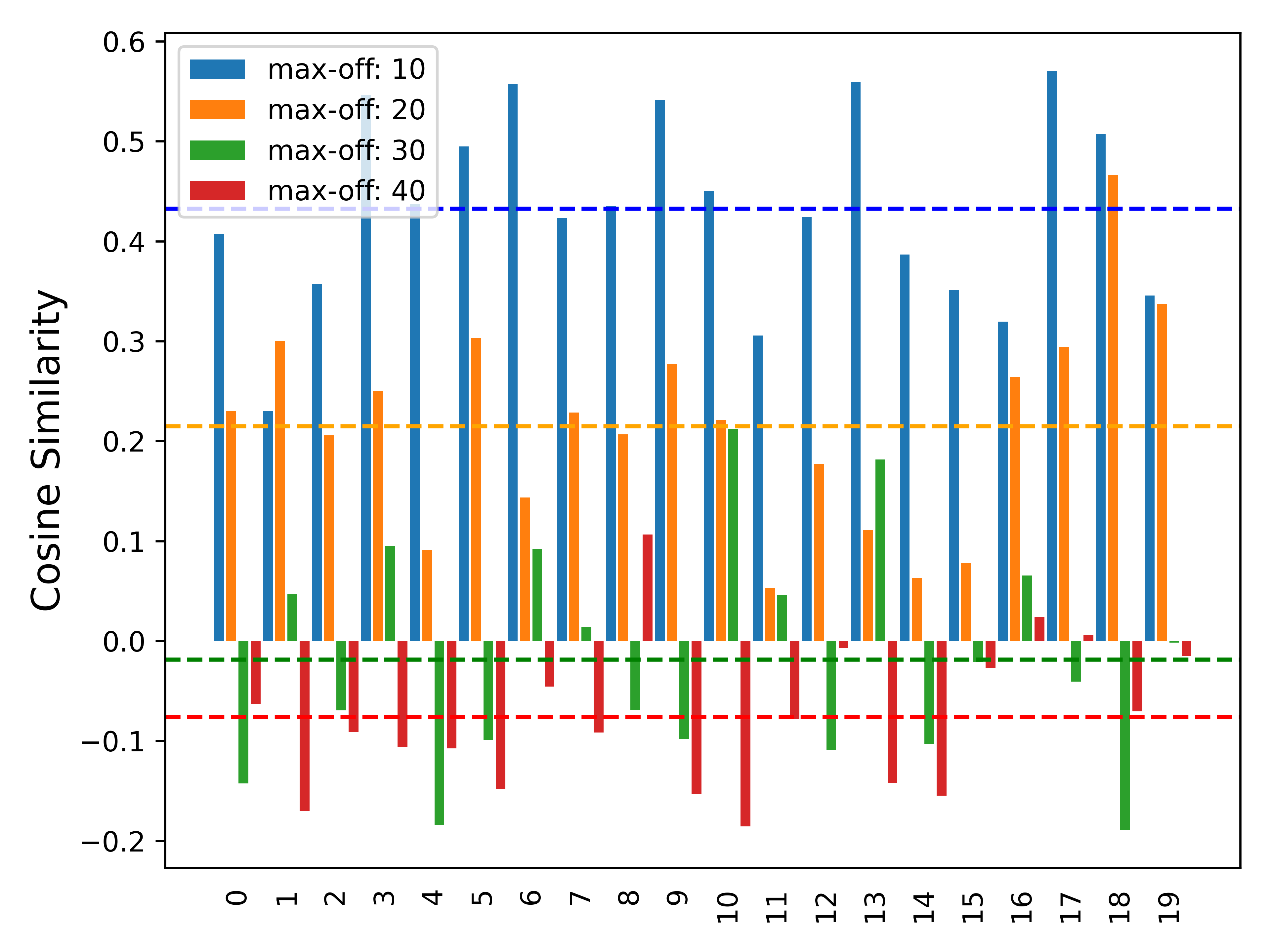}
        \caption{ExFaceGAN(Con)}
        \label{fig:cos_sim_GAN_C_class2}
    \end{subfigure}
	\caption{ Identity-similarity between reference images and samples generated with different max-off values from class positive (Figures \ref{fig:cos_sim_SG2_class1}, \ref{fig:cos_sim_SG3_class1}, and \ref{fig:cos_sim_GAN_C_class1}) and class negative (Figures \ref{fig:cos_sim_SG2_class2}, \ref{fig:cos_sim_SG3_class2}, and \ref{fig:cos_sim_GAN_C_class2}). 
 The horizontal lines represent the mean cosine similarity of the corresponding max-off value. As expected, with larger max-off e.g. 30 and 40, the similarity between the reference and samples from class negative significantly decreased. 
 }
\label{fig:cos_sim_class_1_2}
\vspace{-2mm}
\end{figure}

\textbf{FR Evaluation benchmarks:} We evaluate FR models trained on ExFaceGAN datasets on the following benchmarks: Labeled Faces in the Wild (LFW) \cite{LFW}, AgeDB-30 \cite{agedb}, Celebrities in Frontal to Profile in the Wild (CFP-FP) \cite{cfpfp}, Cross-Age LFW \cite{calfw}, and Cross-Pose LFW \cite{cplfw}. Results for all benchmarks are reported as verification accuracy following their official evaluation protocol.

\begin{table}[h!]
\centering
\resizebox{0.9\linewidth}{!}{%
\begin{tabular}{|c|c|rr|r|}
\hline
\textbf{Class} & \textbf{m} & \textbf{EER$\downarrow$} & \textbf{FMR100$\downarrow$} & \textbf{FDR$\uparrow$} \\ \hline
\multirow{4}{*}{Class positive} & 10 & 0.0012 & 0.0004 & 44.4414 \\
                                & 20 & 0.0045 & 0.0024 & 23.8636 \\
                                & 30 & 0.0127 & 0.0155 & 14.0986 \\
                                & 40 & 0.0293 & 0.0609 & 8.8724  \\ \hline
\multirow{4}{*}{Class negative} & 10 & 0.0014 & 0.0005 & 41.0721 \\
                                & 20 & 0.0070 & 0.0052 & 20.9140 \\
                                & 30 & 0.0187 & 0.0324 & 12.5087 \\
                                & 40 & 0.0375 & 0.1013 & 8.0171  \\ \hline
\multirow{4}{*}{Both classes}   & 10 & 0.0015 & 0.0007 & 41.9367 \\
                                & 20 & 0.0059 & 0.0037 & 22.2112 \\
                                & 30 & 0.0156 & 0.0233 & 13.3946 \\
                                & 40 & 0.0323 & 0.0779 & 8.5926  \\ \hline
\end{tabular}%
}
\caption{Identity verification of samples of class positive, class negative, and both classes, indicating the identity separability in each setting. Data is generated by ExFaceGAN(SG3). m is the max-off value.
}
\label{tab:SG3_class_comparison}
\vspace{-5mm}
\end{table}

\vspace{-3mm}
\section{Results}
\label{sec:results}

\textbf{Identity similarity between references and samples from class positive and class negative:}
Qualitative samples from class negative and positive generated by our  ExFaceGAN(SG2), ExFaceGAN(SG3), and ExFaceGAN(Con) with different max-off values are shown in Figure \ref{fig:samples_dirgan}. Larger intra-class variations can be easily perceived when samples are generated with a large max-off value e.g. 30 or 40. Samples from class positive, as expected, are visually similar to the reference images. On the other hand, the samples from class negative are dissimilar to the reference images, especially, on max-off of 30 and 40. This observation is supported by the results presented in Figure \ref{fig:cos_sim_class_1_2}, where we performed a 1:N comparison between feature representations of a reference and samples from class negative and class positive. 
It can be noticed that increasing the max-off value slightly affects the identity similarity between reference and class positive samples. 
As expected, the similarity between the reference and class negative samples significantly decreases when the max-off is higher than 10, as shown in Figure \ref{fig:cos_sim_class_1_2}. For example, one can notice that samples from class negative in \ref{fig:SG2_intra_class_variation} are of a different gender than the reference and samples from class positive.
This motivated our next experiment to investigate whether class negative samples form a new synthetic identity.


\begin{table}[h!]
\centering
\resizebox{0.9\linewidth}{!}{%
\begin{tabular}{|c|rr|r|}
\hline
\textbf{Model}                   & EER$\downarrow$ & FMR100$\downarrow$ & FDR$\uparrow$ \\ \hline
DiscoFaceGAN \cite{DiscoFaceGAN} & 0.0185          & 0.0304             & 10.3884       \\
GAN-Control \cite{GAN_Control}   & 0.0229          & 0.0468             & 9.1621        \\
SFace \cite{SFace}               & 0.2011          & 0.6315             & 1.3454        \\
InterFaceGAN \cite{InterFaceGAN} & 0.0898          & 0.2565             & 3.3981        \\
DigiFace \cite{digiface}         & 0.0845          & 0.2198             & 3.9365        \\ \hline
ExFaceGAN(SG2)10-Ours            & 0.0011          & 0.0004             & 38.4804       \\
ExFaceGAN(SG2)20-Ours            & 0.0082          & 0.0069             & 16.3796       \\
ExFaceGAN(SG2)30-Ours            & 0.0292          & 0.0627             & 8.4524        \\
ExFaceGAN(SG2)40-Ours            & 0.0689          & 0.2802             & 4.4631        \\ \hline
ExFaceGAN(SG3)10-Ours            & 0.0012          & 0.0004             & 44.4414       \\
ExFaceGAN(SG3)20-Ours            & 0.0045          & 0.0024             & 23.8636       \\
ExFaceGAN(SG3)30-Ours            & 0.0127          & 0.0155             & 14.0986       \\
ExFaceGAN(SG3)40-Ours            & 0.0293          & 0.0609             & 8.8724        \\ \hline
ExFaceGAN(Con)10-Ours            & 0.0214          & 0.0421             & 9.4507        \\
ExFaceGAN(Con)20-Ours            & 0.0221          & 0.0429             & 9.3022        \\
ExFaceGAN(Con)30-Ours            & 0.0311          & 0.0724             & 7.8676        \\
ExFaceGAN(Con)40-Ours            & 0.0542          & 0.1540             & 5.6668        \\ \hline
\end{tabular}%
} 
\caption{Identity verification of SOTA identity-conditional synthetic data and our ExFaceGAN.}
\label{tab:DIRGAN_distributions}
\vspace{-2mm}
\end{table}

\textbf{Identity-discrimination of ExFaceGAN:}
We investigate first the identity discrimination in samples from class positive and class negative generated  by ExFaceGAN. For each of the 5k reference images, we generated 50 class positive samples and 50 class negative samples using our ExFaceGAN(SG3). For each reference image, we consider the class position samples as one identity and the class negative ones as a second identity. Thus, we generated a total of 500k images ($2 \times 5k \times 50$) of 10K identities.
These generated samples formed three testing datasets. The first one contains class positive samples. The second one contains class negative samples and the third dataset contains both, class positive and negative samples.  
Table \ref{tab:SG3_class_comparison} presents the identity verification on the three datasets. We made three main observations: 1) Each of the class positive and class negative samples are, to a very large degree, identity discriminate where for example, EERs are 0.0012 and 0.0014 for class positive and class negative samples, respectively. 2) Increasing the max-off values increased, to a small degree, the verification error rates. 3) Class positive and class negative samples do not (or to extremely a very low degree) share identity information. For example, the EER is 0.0015 when the evaluation dataset contains both class positive and class negative samples, which is slightly higher in comparison to the EER of each of the class positive and class negative scoring 0.0012 and 0.0014, respectively.
This clearly indicates that class negative samples form a new synthetic identity, and thus, our ExFaceGAN can generate two discriminative synthetic identities from each reference image.

\begin{figure*}[h!]
	\centering
    \begin{subfigure}[t]{0.23\linewidth}
        \centering
        \includegraphics[width=\linewidth]{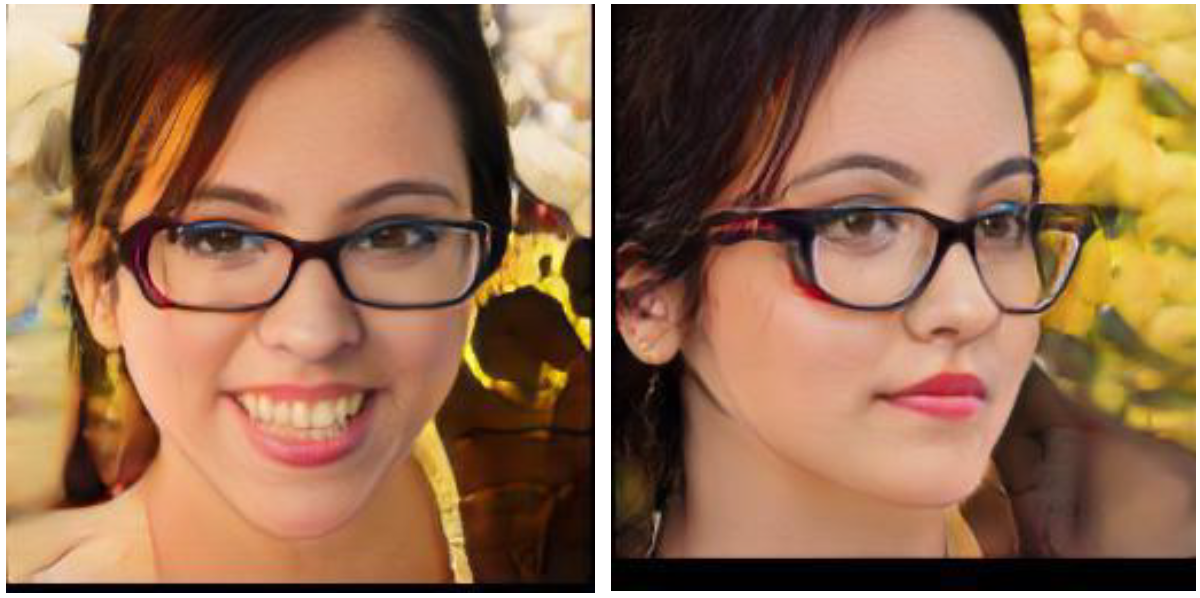}
        \caption{DiscoFaceGAN \cite{DiscoFaceGAN}}
    \end{subfigure}
    \begin{subfigure}[t]{0.23\linewidth}
        \centering
        \includegraphics[width=\linewidth]{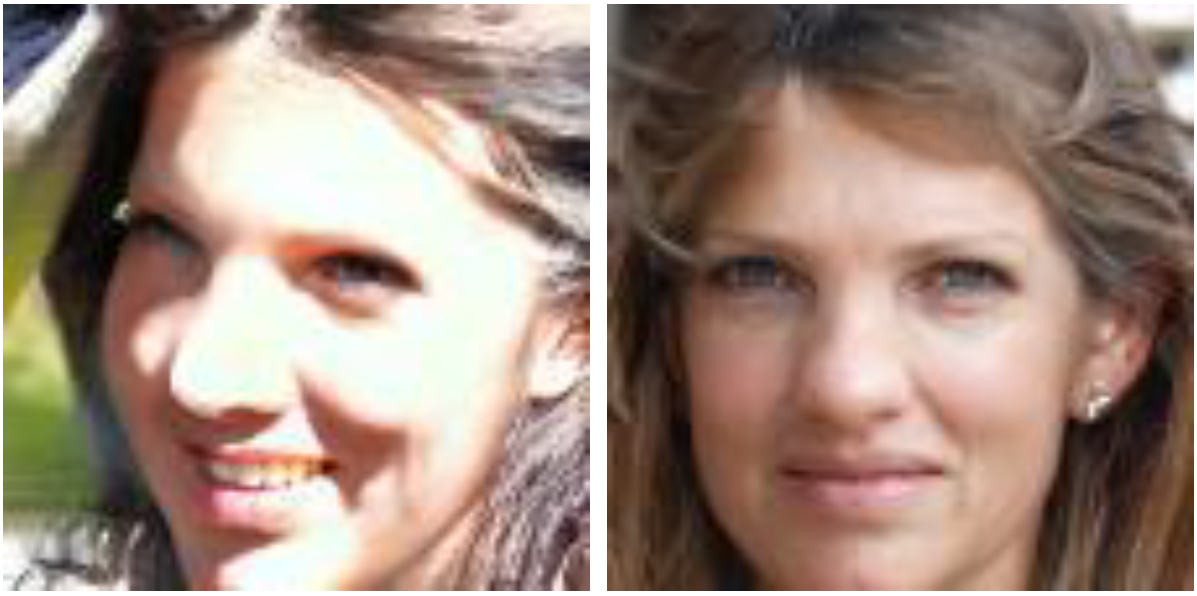}
        \caption{GAN-Control \cite{GAN_Control}}
    \end{subfigure}
    \begin{subfigure}[t]{0.23\linewidth}
        \centering
        \includegraphics[width=\linewidth]{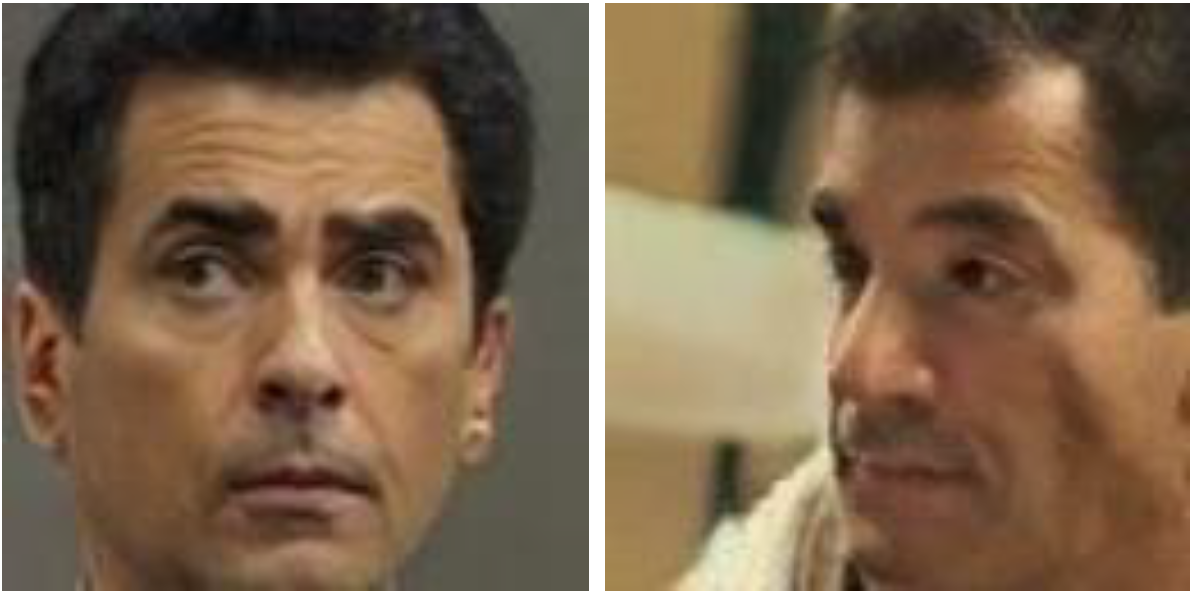}
        \caption{SFace \cite{SFace}}
    \end{subfigure}
    \begin{subfigure}[t]{0.23\linewidth}
        \centering
        \includegraphics[width=\linewidth]{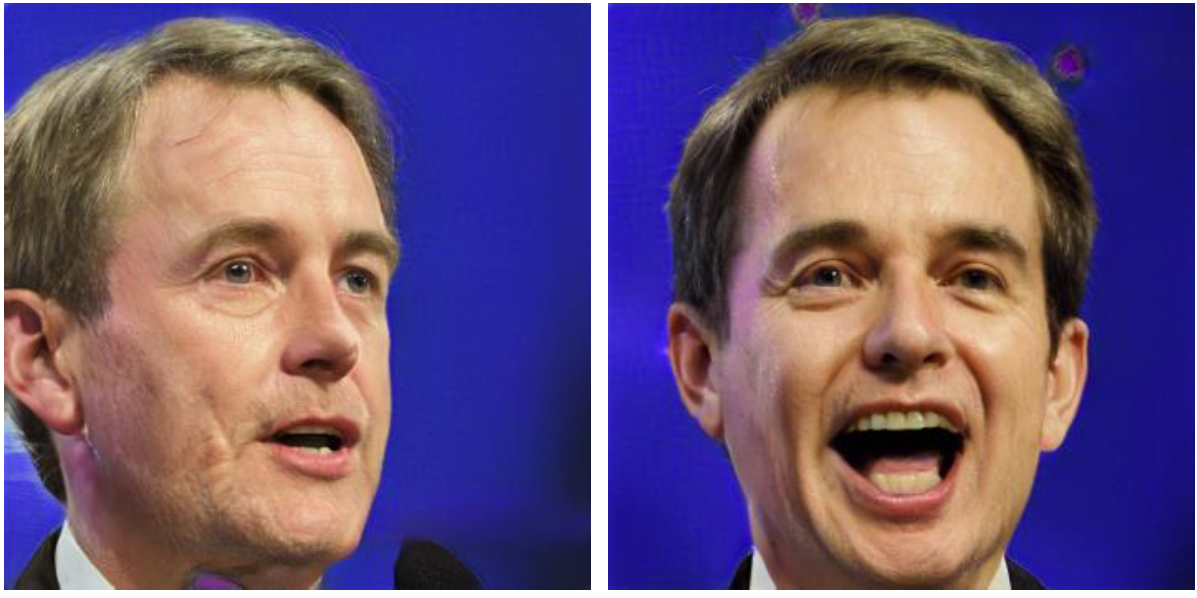}
        \caption{InterFaceGAN \cite{InterFaceGAN}}
    \end{subfigure}
    \begin{subfigure}[t]{0.23\linewidth}
        \centering
        \includegraphics[width=\linewidth]{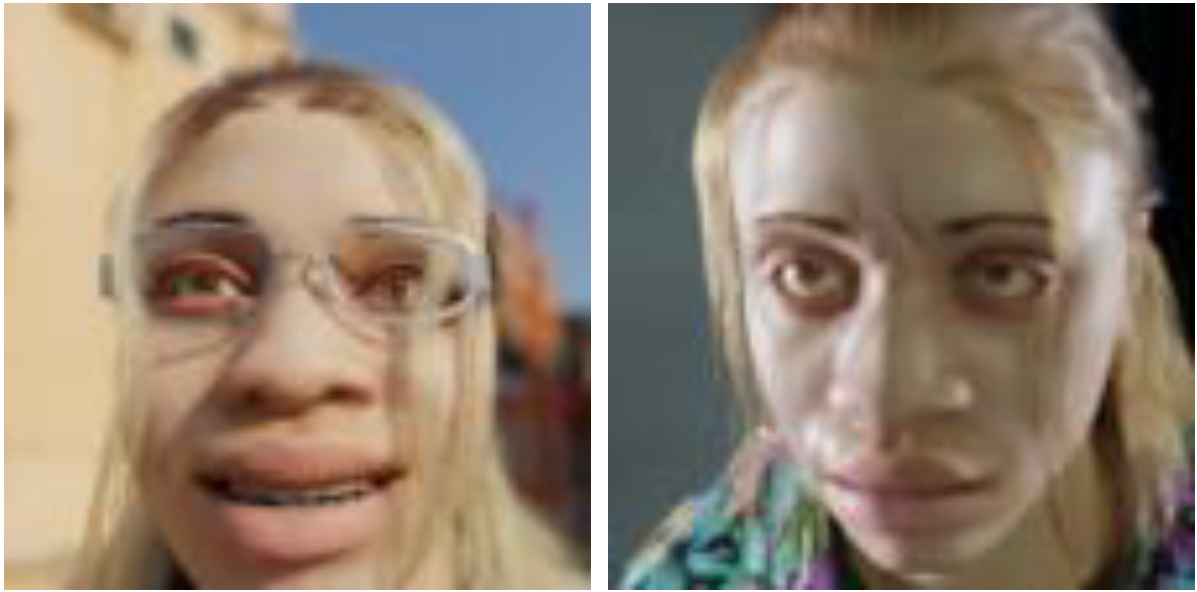}
        \caption{DigiFace \cite{digiface}}
    \end{subfigure}
    \begin{subfigure}[t]{0.23\linewidth}
        \centering
        \includegraphics[width=\linewidth]{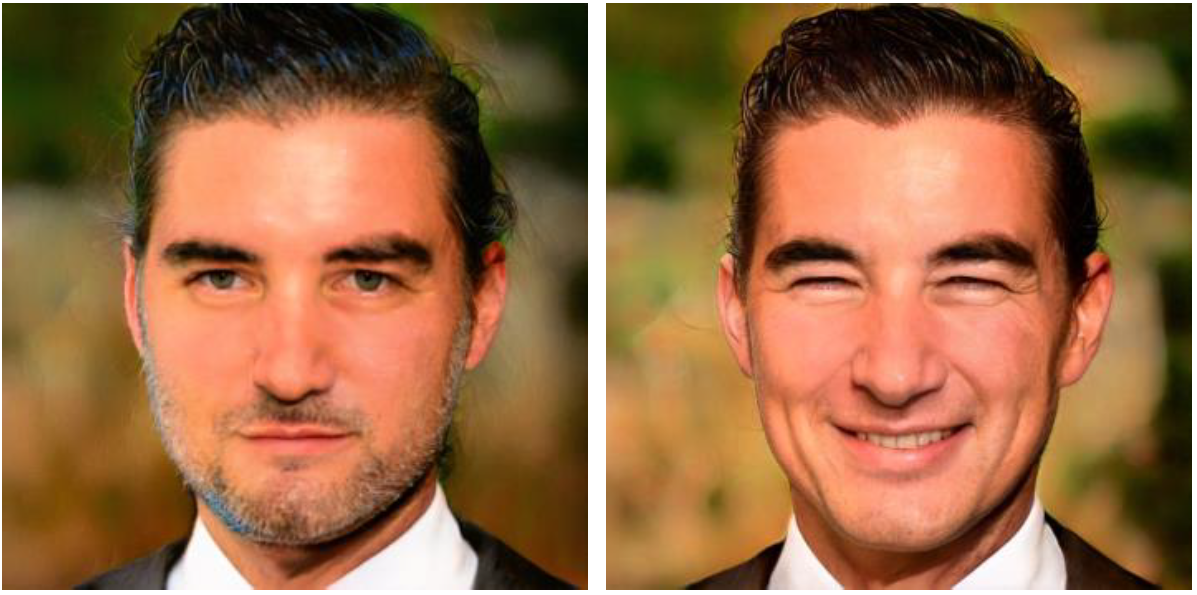}
        \caption{ExFaceGAN(SG2) - Ours}
    \end{subfigure}
    \begin{subfigure}[t]{0.23\linewidth}
        \centering
        \includegraphics[width=\linewidth]{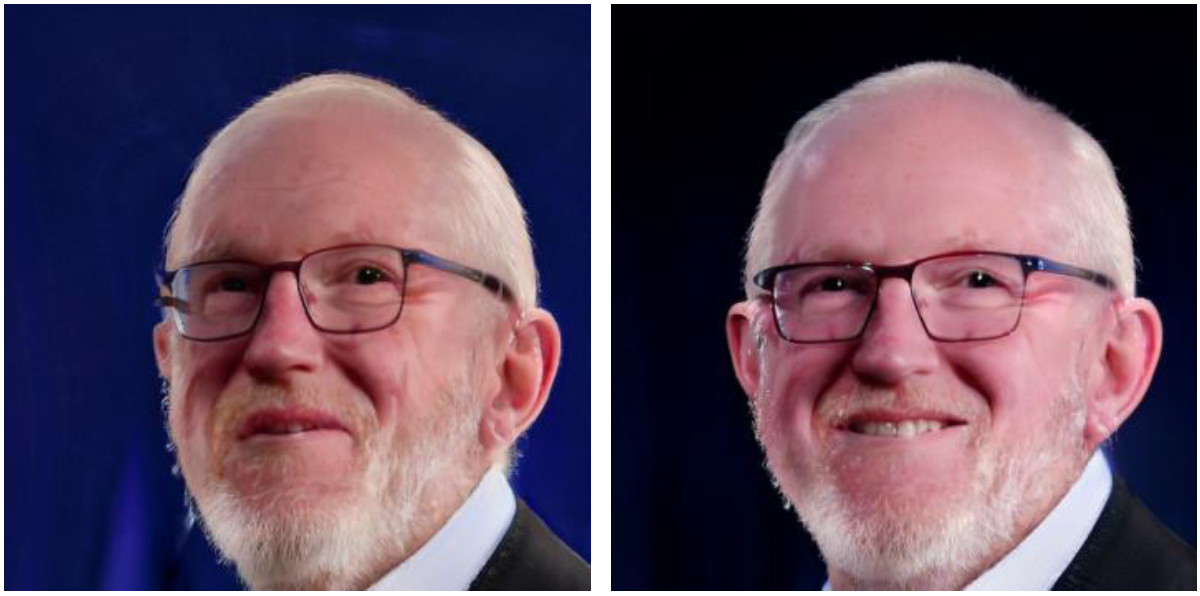}
        \caption{ExFaceGAN(SG3) - Ours}
    \end{subfigure}
    \begin{subfigure}[t]{0.23\linewidth}
        \centering
        \includegraphics[width=\linewidth]{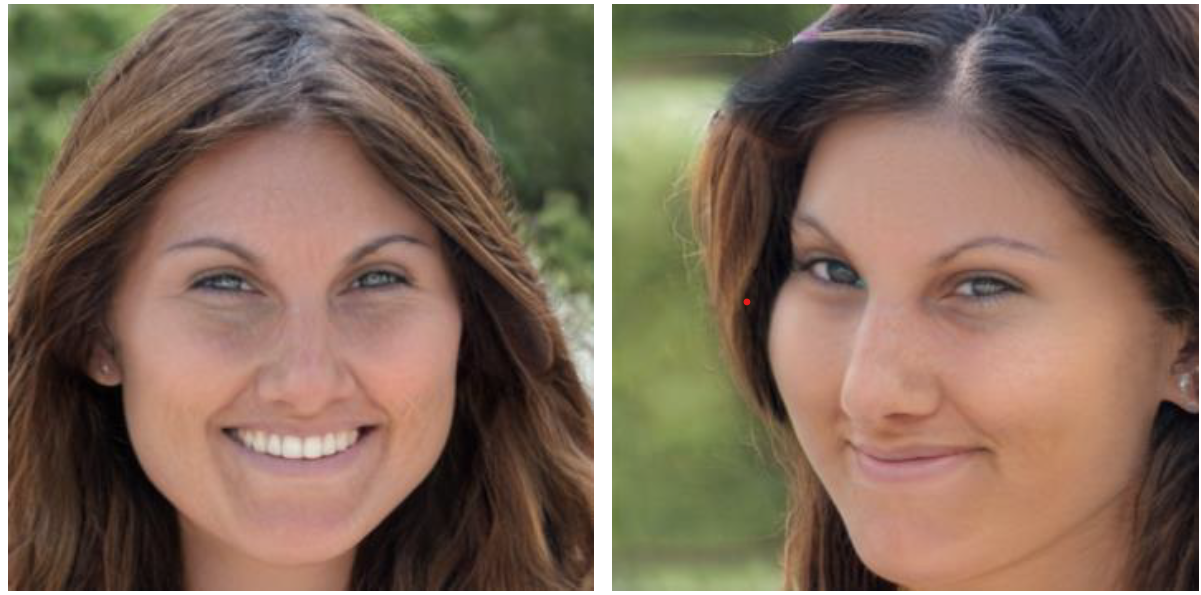}
        \caption{ExFaceGAN(Con) - Ours}
    \end{subfigure}
	\caption{Sample images of different SOTA identity-conditional generative models and our ExFaceGAN.}
	\label{fig:sota_images}
\vspace{-1mm}
\end{figure*}

\textbf{Comparison with SOTA identity-conditioned generative models:} 
We empirically proved the generalizability of our ExFaceGAN approach by applying it to the learned latent spaces of three SOTA GAN models and comparing the identity discrimination of our ExFaceGAN with SOTA approaches, DiscoFaceGAN \cite{DiscoFaceGAN}, GAN-Control \cite{GAN_Control}, SFace \cite{SFace}, InterFaceGAN \cite{InterFaceGAN}, and DigiFace \cite{digiface}. 
Figure \ref{fig:sota_images} presents sample images generated by several SOTA models and our ExFaceGAN. For each approach, we show two images that belong to that same synthetic identity. 

Table \ref{tab:DIRGAN_distributions}
presents the identity verification outcome on synthetic datasets generated by our ExFaceGAN and several SOTA GAN-based models. For each of the ExFaceGAN approaches, we present the results using a max-off of 10, 20, 30, and 40. 
As shown in Table \ref{tab:DIRGAN_distributions}, InterFaceGAN achieved lower identity verification accuracies in comparison to our ExFaceGAN and other SOTA approaches.
Unlike previous identity-conditioned generative models, our ExFaceGAN offers a controllable trade-off between identity discrimination and intra-class variation using the max-off parameter. 
As we discussed in the previous subsection, increasing the max-off value leads to more challenging comparison pairs (higher intra-class variation) which slightly decreases verification accuracies, however, such challenging samples are needed, for example, to train FR as we will present in the next section. 
Unlike DiscoFaceGAN \cite{DiscoFaceGAN} and GAN-Control \cite{GAN_Control}, the intra-class variation in our ExFaceGAN is not limited to a predefined set of attributes. Although, we demonstrated that our ExFaceGAN could be integrated into GAN-Control \cite{GAN_Control}, enhancing identity discrimination.
Our approach, unlike DiscoFaceGAN, GAN-Control, and SFace, does not require designing or training a special architecture. Our ExFaceGAN(Con) achieved lower verification accuracies than our ExFaceGAN(SG2) and ExFaceGAN(SG3). However, combining attributes conditions of GAN-Control with our ExFaceGAN created more challenging pairs with large intra-class variations as shown in Figure \ref{fig:samples_dirgan}, which is beneficial for application use cases such as training FR using synthetic data as we demonstrate in the next section.

\begin{table}[h!]
\centering
\resizebox{\linewidth}{!}{%
\begin{tabular}{|c|c|r|c|c|c|c|c|}
\hline
\textbf{Model}                  & \textbf{m} & \textbf{LFW}   & \textbf{AgeDB-30} & \textbf{CFP-FP} & \textbf{CA-LFW} & \textbf{CP-LFW} & \textbf{BC} \\ \hline
\multirow{4}{*}{ExFaceGAN(SG2)} & 10 & 79.88 & 52.82          & 61.83          & 62.97 & 60.12          & 4  \\
                                & 20 & 82.47 & 56.32          & 63.94          & 67.18 & 62.40          & 21 \\
                                & 30 & 82.05 & 56.48          & 66.97          & 66.85 & 61.18          & 21 \\
                                & 40 & 81.67 & 55.22          & 66.97          & 65.48 & 61.12          & 15 \\ \hline
\multirow{4}{*}{ExFaceGAN(SG3)} & 10 & 78.27 & 51.20          & 62.27          & 61.87 & 59.77          & 1  \\
                                & 20 & 84.10 & 53.28          & 66.41          & 66.35 & 64.47          & 20 \\
                                & 30 & 84.37 & 54.82          & 68.46          & 68.67 & 65.00          & 30 \\
                                & 40 & 84.05 & 56.22          & 69.10          & 68.42 & 64.45          & 28 \\ \hline
\multirow{4}{*}{ExFaceGAN(Con)} & 10               & \textbf{87.83} & 57.17             & 69.33           & \textbf{72.30}  & 67.38           & \textbf{52} \\
                                & 20 & 87.68 & 56.67          & 69.17          & 71.53 & \textbf{67.58} & 48 \\
                                & 30 & 86.57 & \textbf{57.27} & 69.17          & 70.07 & 66.92          & 47 \\
                                & 40 & 85.95 & 56.58          & \textbf{70.87} & 69.95 & 66.43          & 43 \\ \hline
\end{tabular}%
}
\caption{Verification accuracies (\%) of FR models trained with 250K images from our ExFaceGAN under different settings. m is the max-off parameter value}
\label{tab:SG_FR_performance}
\vspace{-2mm}
\end{table}

\begin{table}[h!]
\centering
\resizebox{\linewidth}{!}{%
\begin{tabular}{|c|c|c|c|c|c|c|c|}
\hline
\textbf{Model}                          & \textbf{m}    & \textbf{Aug.} & \textbf{LFW}   & \textbf{AgeDB-30} & \textbf{CFP-FP} & \textbf{CA-LFW} & \textbf{CP-LFW} \\ \hline
\multirow{2}{*}{ExFaceGAN(SG2)}            & \multirow{2}{*}{20} & -                          & 82.47          & 56.32             & 63.94           & 67.18           & 62.40           \\
                                        &                     & RA                          & 83.53          & 65.13             & 66.43           & 70.80           & 62.80           \\ \hline
\multirow{2}{*}{ExFaceGAN(SG3)}            & \multirow{2}{*}{30} & -                        & 84.37          & 54.82             & 68.46           & 68.67           & 65.00           \\
                                        &                     & RA                       & 87.13          & 68.98             & 70.67           & 73.90           & 65.83           \\ \hline
\multirow{2}{*}{ExFaceGAN(Con)}         & \multirow{2}{*}{10} & -                            & 87.83          & 57.17             & 69.33           & 72.30           & 67.38           \\
                                        &                     & RA                            & 93.20          & 78.38             & 75.11           & 81.70           & 71.47           \\ \hline
\end{tabular} %
}
\caption{Impact of data augmentation on verification accuracies (\%) of FR models trained with 250K from our ExFaceGAN. m is the max-off parameter value.}
\label{tab:DIRGAN_plus_ra}
\vspace{-1mm}
\end{table}

\begin{table}[h]
\centering
\resizebox{\linewidth}{!}{%
\begin{tabular}{|c|c|c|c|c|c|}
\hline
\textbf{Loss}  & \textbf{LFW}   & \textbf{AgeDB-30} & \textbf{CFP-FP} & \textbf{CA-LFW} & \textbf{CP-LFW} \\ \hline
 ArcFace \cite{ArcFace}       & 89.45          & 71.32             & \textbf{72.86}  & 77.18           & 68.65           \\   \hline
AdaFace \cite{AdaFace}    & 89.62          & 71.80             & 70.94           & 77.37           & 68.42           \\  \hline
 CosFace \cite{CosFace}                     & 90.47          & \textbf{72.85}    & 72.70           & \textbf{78.60}  & \textbf{69.27}  \\  \hline 
ElasticFace-Cos \cite{ElasticFace}         & \textbf{90.62} & 72.17             & 72.07           & 78.42           & 69.08           \\ \hline
\end{tabular}
} %
\caption{Verification accuracies (\%) of FR models trained with different loss functions on 500K images from ExFaceGAN(SG3). 
}
\label{tab:DIRGAN_losses}
\vspace{-4mm}
\end{table}

\textbf{Synthetic-based FR}
We demonstrate that data generated by our ExFaceGAN can be successfully used to train FR. We first conducted an ablation study by generating 12 datasets, each containing 250K images from ExFaceGAN(SG2), ExFaceGAN(SG3), and ExFaceGAN(Con) with different max-off values of 10, 20, 30, and 40. These datasets are used to train FR models with ResNet-50 network architecture \cite{ResNet} and ArcFace loss \cite{ArcFace}, using exact experimental setups described in \cite{ArcFace} and Section \ref{sec:exp_setup}. 
For each ExFaceGAN, the overall comparison of the verification performance is based on the sum of the performance ranking Borda count (BC) on the considered evaluation datasets, following the comparison method in \cite{USynthFace}.  As shown in Table \ref{tab:SG_FR_performance}, ExFaceGAN(SG3) with a max-off of 30 achieved higher verification accuracies than ExFaceGAN(SG2). ExFaceGAN(Con) with a max-off of 10 achieved the highest verification accuracies among all ExFaceGAN approaches.
It should be noted that all results are reported without applying data augmentation during the training. 

\begin{table*}[!ht]
\centering
\resizebox{0.98\linewidth}{!}{%
\begin{tabular}{|c|c|c|c|c|c|c|c|}
\hline
\textbf{Method} & \textbf{Id/Img.} & \textbf{LFW} & \textbf{AgeDB-30} & \textbf{CFP-FP} & \textbf{CA-LFW} & \textbf{CP-LFW} & \textbf{Avg} \\ \hline
CASIA-WebFace \cite{CASIA}           & 10K/46K      & 99.55              & 94.55                 & 95.31                & 93.78                & 89.95  & 94.63   \\ \hline \hline
SynFace \cite{SynFace}                & 10K/50    & 88.98             & -                 & -                 & -                 & -              & -                 \\
SynFace (w/IM) \cite{SynFace}         & 10K/50    & 91.93             & 61.63             & 75.03             & 74.73             & 70.43          & 74.75             \\ \hline
SFace \cite{SFace}                           & 10K/60 & 91.87             & 71.68             & 73.86             & 77.93             & 73.20          & 77.71             \\ \hline
DigiFace-1M \cite{digiface}           & 10K/50    & \textbf{95.40}    & 76.97             & \textbf{87.40}    & 78.62             & \textbf{78.87} & \textbf{83.45}    \\ \hline
USynthFace \cite{USynthFace}                 & 400K/1    & 92.23             & 71.62             & \underline{78.56} & 77.05             & 72.03          & 78.30             \\ \hline
IDnet \cite{Kolf_2023_CVPR}          & 10K/50                                 & 92.58              & 73.53                 & 75.40                & 79.90            & \underline{ 74.25 }               & 79.13    \\ \hline
GAN-Control \cite{GAN_Control} - Ours & 10K/50    & 93.22             & \underline{77.60} & 73.03             & \underline{82.25} & 70.80          & 79.38             \\ \hline
ExFaceGAN(SG3) - Ours                 & 10K/50    & 90.47             & 72.85             & 72.70             & 78.60             & 69.27          & 76.78             \\ \hline
ExFaceGAN(Con) - Ours                 & 10K/50    & \underline{93.50} & \textbf{78.92}    & 73.84             & \textbf{82.98}    & 71.60          & \underline{80.17} \\ \hline
\end{tabular}%
}
\caption{Verification accuracies (\%) on five FR benchmarks achieved by the SOTA synthetic-based FR and our ExFaceGAN(SG3) and ExFaceGAN(Con). The result in the first row is reported using the FR model trained on the authentic dataset as an indication of the performance of an FR model trained on the authentic dataset. 
Best values are marked in bold and the second one is in italics. 
}
\label{tab:comparison_DIRGAN_FR_SOTA}
\vspace{-4mm}
\end{table*}

Motivated by the recent works that proposed synthetic-based FR \cite{USynthFace,digiface}, we demonstrate the effectiveness of introducing data augmentation to FR model training. We utilized rand-augmentation \cite{Randaugment} with settings presented by \cite{USynthFace}. Table \ref{tab:DIRGAN_plus_ra} presents the verification accuracies of FR models trained with our ExFaceGAN using the best max-off values from Table \ref{tab:SG_FR_performance}. It can be clearly observed that applying data augmentation consistently improves the verification accuracies for all ExFaceGAN models.

\textbf{Comparison with SOTA synthetic-based FR}
We compare ExFaceGAN with the recent SOTA  synthetic-based FR models. 
To provide a fair comparison, we generate 500K images of 10K identities, each containing 50 images. The images are generated from both, class positive and class negative. 
We first provide an ablation study on different training loss functions, including ArcFace \cite{ArcFace}, AdaFace \cite{AdaFace}, CosFace \cite{CosFace} and Elastic-CosFace \cite{ElasticFace}, as shown in Table \ref{tab:DIRGAN_losses}. It can be observed that CosFace achieved the best overall verification accuracies, followed by Elastic-CosFace.
Table \ref{tab:comparison_DIRGAN_FR_SOTA} presents comparisons of our ExFaceGAN(SG3) and ExFaceGAN(Con) (trained using CosFace loss) with SOTA synthetic-based FR approaches. 
Our ExFaceGAN approaches achieved the best verification accuracies on Cross-Age datasets (AgeDB-30 and CA-LFW). Also, ExFaceGAN achieved very competitive performance to SOTA FR models trained with synthetic data on CFP-FP and CP-LFW. On LFW, our ExFaceGAN outperformed FR models trained with data generated by GAN, including SynFace, SFace, IDNet, and UsynthFace, and scored behind DigiFace-1M which is based on a computationally expensive digital rendering pipeline.  Results of SOTA approaches in Table \ref{tab:comparison_DIRGAN_FR_SOTA} are reported as in their  corresponding works. We opted to train and evaluate an FR model on GAN-Control \cite{GAN_Control} to provide a direct comparison with our ExFaceGAN(Con). Our ExFaceGAN(Con) outperformed GAN-Control \cite{GAN_Control} on the considered benchmarks.

\vspace{-2mm}
\section{Conclusion}
\label{sec:conclusion}
\vspace{-1mm}
We proposed in this work a framework, ExFaceGAN, to disentangle complex identity information in the learned latent space of StyleGAN models.
The proposed ExFaceGAN enabled the generation of multiple samples of specific synthetic identity without the need to design and train a dedicated deep generative model or supervision from attribute classifiers. We also proposed a controllable sampling technique to gain control over the balance between the intra-class variations and inter-class compactness in our generated data. 
We empirically proved the generalizability of the effectivity of our framework by integrating it in learned latent spaces of three SOTA GAN models, StyleGAN-ADA, StyleGAN-3, and GAN-Control. We also demonstrated that the generated data by ExFaceGAN can be successfully used to train FR models, advancing SOTA performances on a number of benchmarks for synthetic-based FR.

\textbf{Acknowledgment}
This research work has been funded by the German Federal Ministry of Education and Research and the Hessian Ministry of Higher Education, Research, Science and the Arts within their joint support of the National Research Center for Applied Cybersecurity ATHENE. This work has been partially funded by the German Federal Ministry of Education and Research (BMBF) through the Software Campus Project.

{\small
\bibliographystyle{ieee}
\bibliography{main}
}

\end{document}